\documentclass{article} 
\usepackage{iclr2026_conference,times}

\usepackage{amsmath,amsfonts,bm}

\def\eqref#1{equation~\ref{#1}}

\def\1{\bm{1}}

\DeclareMathAlphabet{\mathsfit}{\encodingdefault}{\sfdefault}{m}{sl}
\SetMathAlphabet{\mathsfit}{bold}{\encodingdefault}{\sfdefault}{bx}{n}

\usepackage{hyperref}
\usepackage{booktabs}       
\usepackage{url}
\usepackage{amsmath}
\usepackage{amsfonts} 
\usepackage{multirow}
\usepackage[capitalise,noabbrev]{cleveref}
\usepackage{graphicx}  
\usepackage{subcaption}  
\title{Look-ahead Reasoning with a Learned Model\\ in Imperfect Information Games}

\author{Ondřej Kubíček \\
Artificial Intelligence Center, FEE\\
Czech Technical University in Prague \\
Czech Republic\\ 
\texttt{kubicon3@fel.cvut.cz}
\And 
Viliam Lisý \\
Artificial Intelligence Center, FEE\\
Czech Technical University in Prague \\
Czech Republic\\ 
\texttt{viliam.lisy@agents.fel.cvut.cz}
} 

\iclrfinalcopy 
\begin{document}

\newcommand{\RealNumbers}[0]{\mathbb{R}}
\newcommand{\BinaryNumbers}[0]{\mathbb{B}}
\newcommand{\NaturalNumbers}[0]{\mathbb{N}}
\newcommand{\Expectation}[0]{\mathbb{E}}

\newcommand{\Simplex}{\Delta}

\newcommand{\FOSG}[0]{\mathcal{G}}

\newcommand{\PublicIndex}[0]{0}
\newcommand{\InitialIndex}[0]{\text{INIT}}
\newcommand{\BRIndex}[0]{BR}

\newcommand{\Players}[0]{\mathcal{N}}
\newcommand{\Player}[0]{i}
\newcommand{\OtherPlayer}[0]{j}
\newcommand{\LastPlayer}[0]{N}
\newcommand{\PlayerFunction}[0]{p}

\newcommand{\WorldStates}[0]{\mathcal{W}}
\newcommand{\WorldState}[0]{w}
\newcommand{\InitWorldState}[0]{\WorldState^{\InitialIndex}}

\newcommand{\Histories}[0]{\mathcal{H}}
\newcommand{\History}[0]{h}
\newcommand{\InitHistory}[0]{\History^{\InitialIndex}}
\newcommand{\HistoryExtend}{\sqsubseteq}

\newcommand{\Actions}[1]{\mathcal{A}_{#1}}
\newcommand{\Action}[1]{a_{#1}}
\newcommand{\ActionLogit}[1]{A_{#1}}

\newcommand{\Observations}[1]{\mathcal{O}_{#1}}
\newcommand{\Observation}[1]{o_{#1}}
\newcommand{\ObservationSet}[1]{\mathbb{O}_{#1}}
\newcommand{\PublicObservations}[0]{\Observations{\PublicIndex}}
\newcommand{\PublicObservation}[0]{\Observation{\PublicIndex}}
\newcommand{\PublicObservationSet}[0]{\ObservationSet{\PublicIndex}}

\newcommand{\Rewards}[1]{\mathcal{R}_{#1}}
\newcommand{\Reward}[1]{r_{#1}}

\newcommand{\Utility}[1]{u_{#1}}

\newcommand{\Transitions}[0]{\mathcal{T}}

\newcommand{\Infosets}[1]{\mathcal{S}_{#1}}
\newcommand{\Infoset}[1]{s_{#1}}
\newcommand{\PublicStates}[0]{\Infosets{\PublicIndex}}
\newcommand{\PublicState}[0]{\Infoset{\PublicIndex}}
\newcommand{\Abstracted}[1]{\overline{#1}}

\newcommand{\Trajectory}[0]{\tau}
\newcommand{\TrajectoryLength}[0]{l}
\newcommand{\TrajectoryStep}[0]{t}

\newcommand{\TimeStep}[0]{t}
\newcommand{\UnrollStep}[0]{k}
  
\newcommand{\Strategy}[1]{\pi_{#1}}
\newcommand{\Strategies}[1]{\Pi_{#1}}
\newcommand{\Nash}[1]{\Strategy{#1}^*}
\newcommand{\EpsilonNash}[1]{\Strategy{#1}^{\epsilon}}
\newcommand{\BRStrategy}[1]{\Strategy{#1}^{\BRIndex}}
\newcommand{\BRSet}[1]{\BRIndex_{#1}}
\newcommand{\StrategyMapping}[0]{f}

\newcommand{\StrategyUtility}[2]{\Utility{#1}^{#2}}
\newcommand{\CounterfactualValue}[2]{v_{#1}^{#2}}

\newcommand{\Exploitability}[0]{\mathcal{E}}

\newcommand{\Reach}[1]{P^{#1}}
\newcommand{\PlayerReach}[2]{\Reach{#2}_{#1}}

\newcommand{\AbstractionFunction}[0]{\Lambda}
\newcommand{\LegalActionsFunction}[0]{\Gamma}
\newcommand{\RepresentationFunction}[0]{\AbstractionFunction^{I}}
\newcommand{\DynamicsFunction}[0]{\Upsilon}
\newcommand{\PropertiesFunction}[0]{\kappa}
\newcommand{\DistanceFunction}[0]{d}
\newcommand{\StrategyFunction}[0]{\pi}  
\newcommand{\TransformationsFunction}[0]{\tau}
\newcommand{\ValueFunction}[0]{v}
\newcommand{\DecoderFunction}[0]{\AbstractionFunction^{-1}}

\newcommand{\InfosetProperty}[1]{\phi_{#1}}
\newcommand{\SoftmaxWeight}[0]{\omega}
\newcommand{\SoftmaxTemperature}[0]{\gamma}

\newcommand{\NeuralParameters}[0]{\theta}
\newcommand{\NeuralFunction}[1]{#1_\NeuralParameters}

\newcommand{\AbstractionModel}[0]{\NeuralFunction{\AbstractionFunction}}
\newcommand{\LegalActionsModel}[0]{\NeuralFunction{\LegalActionsFunction}}
\newcommand{\RepresentationModel}[0]{\NeuralFunction{\RepresentationFunction}}
\newcommand{\DyanmicsModel}[0]{\NeuralFunction{\DynamicsFunction}} 
\newcommand{\PropertiesModel}[0]{\NeuralFunction{\PropertiesFunction}}

\newcommand{\InfosetModel}[0]{\zeta}

\newcommand{\AbstractionLimit}[0]{L}

\newcommand{\LegalThreshold}[0]{\varphi}

\newcommand{\TerminalSignal}[0]{z}
\newcommand{\TerminalLogit}[0]{l}

\newcommand{\Transformation}[0]{\chi}
\newcommand{\Transformations}[0]{T}

\newcommand{\Loss}[0]{\mathcal{L}}
\newcommand{\ModelLoss}[0]{\Loss_\NeuralParameters^{M}}
\newcommand{\ModelAbstractionLoss}[0]{\Loss_\NeuralParameters^{MA}}
\newcommand{\AbstractionLoss}[0]{\Loss_\NeuralParameters^{A}}
\newcommand{\ClusterLoss}[0]{\Loss_\NeuralParameters^C}
\newcommand{\RepresentationLoss}[0]{\Loss_\NeuralParameters^R}
\newcommand{\DynamicsLoss}[0]{\Loss_\NeuralParameters^D}
\newcommand{\LegalActionsLoss}[0]{\Loss_\NeuralParameters^L}
\newcommand{\TerminalLoss}[0]{\Loss_\NeuralParameters^T}
\newcommand{\RewardLoss}[0]{\Loss_\NeuralParameters^R} 
\newcommand{\InfosetLoss}[0]{\Loss_\NeuralParameters^{CE}}
\newcommand{\InfosetModelLoss}[0]{\Loss_\NeuralParameters^{S}}
\newcommand{\PolicyGradientLoss}[0]{\Loss_\NeuralParameters^{PG}}
\newcommand{\ValueFunctionLoss}[0]{\Loss_\NeuralParameters^{V}}
\newcommand{\TransformationsLoss}[0]{\Loss_\NeuralParameters^{T}}
\newcommand{\NeurdLoss}[0]{\Loss_\NeuralParameters^{N}}
\newcommand{\RepulsionLoss}[0]{\Loss_\NeuralParameters^{\text{Rep}}}
\newcommand{\HardLoss}[0]{\Loss_\NeuralParameters^{\text{Hard}}}

\newcommand{\Distance}[0]{\mathcal{D}}

\newcommand{\ActionValue}[1]{Q_{#1}}
\newcommand{\StateValue}[1]{V_{#1}}
\newcommand{\Regret}[1]{\mathcal{R}_{#1}}

\newcommand{\StrategyDirection}[0]{\delta}

\newcommand{\AlgorithmName}[0]{LAMIR}

\newcommand{\RepulsionConstant}[0]{d_r}
\newcommand{\HardClusteringDist}[0]{d_h}
\newcommand{\SoftClusteringDist}[0]{d_s}

\newcommand{\DepthLimit}[0]{D}
\newcommand{\DepthIterator}[0]{d}

\maketitle

\begin{abstract}
  Test-time reasoning significantly enhances pre-trained AI agents' performance. However, it requires an explicit environment model, often unavailable or overly complex in real-world scenarios. While MuZero enables effective model learning for search in perfect information games, extending this paradigm to imperfect information games presents substantial challenges due to more nuanced look-ahead reasoning techniques and large number of states relevant for individual decisions. This paper introduces an algorithm \AlgorithmName{} that learns an abstracted model of an imperfect information game directly from the agent-environment interaction. During test time, this trained model is used to perform look-ahead reasoning. The learned abstraction limits the size of each subgame to a manageable size, making theoretically principled look-ahead reasoning tractable even in games where previous methods could not scale. We empirically demonstrate that with sufficient capacity, \AlgorithmName{} learns the exact underlying game structure, and with limited capacity, it still learns a valuable abstraction, which improves game playing performance of the pre-trained agents even in large games.
\end{abstract}

\section{Introduction}
\label{sec:introduction}
Strategic reasoning and planning are key components of human intelligence, encompassing our ability to reason about possible outcomes of actions in complex situations, often with incomplete information and uncertain consequences. Although humans navigate such decision-making naturally, replicating this process in artificial intelligence remains a fundamental challenge. Games, with their well-defined rules and yet complex strategic landscapes, serve as ideal testbeds for developing and evaluating AI planning and reasoning methods \citep{mnih2015,alphazero2018,stratego2022,llmboardgames2025}.

In perfect information games like Chess, Go or Shogi, look-ahead search algorithms as Minimax and Monte Carlo Tree Search (MCTS) have achieved superhuman performance by systematically exploring future states \citep{russell2016,alphazero2018}. These methods traditionally rely on access to game rules to implement state transitions in the search. MuZero demonstrated that an agent can learn a model of the environment dynamics implicitly through interaction and use this learned model to perform MCTS planning, removing the dependency on explicit implementation of the rules \citep{muzero2020}.

Not requiring explicit pre-programmed representation of the environment greatly expands applicability of AI methods. A method that does not require explicit rules representation can be applied, for example, to create an AI opponent in a proprietary video game without access to its source code; to create AI opponents for a large database of games for an online game playing platform, where programming a suitable representation for each of them would be prohibitively expensive; or in a game design setting, where the game is repetitively modified without the need for a programmer to reflect the changes in the implementation. 

However, extending the model learning paradigm to imperfect information games such as Poker or Stratego presents fundamental difficulties. Since players lack complete knowledge of the state of the game, theoretically sound look-ahead reasoning methods need to reason about the distribution of all possible hidden states consistent with shared knowledge \citep{valuefunctions2023}, which differs from the MCTS used by MuZero.


Our work aims to \textbf{enable look-ahead reasoning in two-player imperfect information games using a learned abstract model}, thereby \textbf{eliminating the need for explicit game rules} and also \textbf{enabling look-ahead reasoning in parts of the game intractable without abstraction}. Following the approach of \citet{muzero2020}, we focus only on games without chance events, which is a large class that includes commonly used benchmarks like Dark Chess, Stratego, Battleship or Imperfect Information Goofspiel. This allows us to tackle the unique difficulties of learning an effective abstraction for imperfect information without conflating it with the separate challenges introduced by chance events. Our contributions are: We identify the necessary components that a learned model must capture to facilitate look-ahead reasoning under imperfect information. We introduce a training procedure to learn these components from sampled game trajectories. We demonstrate how tractably small, domain-independent abstractions can be learned concurrently with the model. Finally, we introduce a way to conduct look-ahead reasoning with the learned model.

Our empirical evaluation shows that in small games, given sufficient capacity, the strategies produced by look-ahead reasoning are less exploitable than those of concurrently trained Regularized Nash Dynamics (RNaD). Furthermore, in large games with shared knowledge consistent with over $10^{18}$ states for some decisions, the proposed look-ahead reasoning is still applicable in all decision points and significantly improves over RNaD, reaching up to 80\% win rate in head-to-head play.

\subsection{Related Work}
\paragraph{Direct Policy Optimization} One approach for computing strategies in large imperfect information games stores the strategy implicitly in neural network weights and directly optimizes this policy based on self-play traces. These methods include: policy-gradient algorithms with reward regularization, like Regularized Nash Dynamics \citep{poincare2021,stratego2022,sokota2023,perturbation2025}; training networks to approximate CFR results, like Deep CFR or DREAM \citep{brown2019deep,brown2020rebel,steinberger2020dream}; or iteratively training best responses to growing strategy pools, like PSRO \citep{psro2017,alphastar2019}. Critically, these approaches rely solely on the trained actor during gameplay and cannot refine decisions with additional test-time computation. 
Pre-trained agents without test-time reasoning are usually very exploitable~\citep{wang2023adversarial,lisy2016equilibrium}, and adding test-time reasoning greatly improves their capabilities in games~\citep{alphago2016,sepot2024}, and beyond~\citep{snell2024scaling}.
This paper enables adding test-time reasoning to policies created by direct policy optimization algorithms.

\paragraph{Look-ahead reasoning} Reasoning algorithms in imperfect information games, such as Counterfactual Regret Minimization \citep{cfr2007}, iteratively improve player's policies by systematically iterating over all possible future action sequences in all possible (unobserved) states of the game. In large games, this requires either domain-specific abstractions, like Libratus \citep{libratus2018} or depth-limited reasoning, like DeepStack, ReBeL, Student of Games, SePoT \citep{deepstack2017, brown2020rebel, studentofgames2023, sepot2024}. In either case, all these algorithms require explicit implementation of game rules to construct game trees and manage belief states. It limits their applicability when exact rules are unavailable, computationally prohibitive or if the amount of possible hidden states is too huge. Knowledge-limited subgame solving \citep{zhang2021klss, zhang2025kluss, safeklss2023} can reduce the complexity, but even this reduced state space remains intractable in the games we study here. In contrast, our approach does not require explicit game rules and automatically learns a tractably small abstraction of the game just from full traces of game play.


\paragraph{Model learning} In single-player settings, Dreamer and subsequent works showed that learning models and generating artificial traces can match purely model-free approaches \citep{dreamer2020,dreamer2021,dreamer2025}. MuZero demonstrated similar results in perfect information games, using learned models for reasoning during gameplay \citep{muzero2020,stochasticmuzero2022}. Our work extends these approaches to imperfect information games without chance.

\section{Background}
\label{sec:background}
We define two-player zero-sum simultaneous move game as $\FOSG =  (\Players, \WorldStates, \InitWorldState, \Actions{}, \Transitions, \Rewards{},  \Observations{})$ \cite{fosg2022}, where $\Players = \{1, 2\}$ are the players, $\WorldStates$ is the set of world states in the game and $\InitWorldState \in \WorldStates$ is the initial world state. $\Actions{} = \Pi_{\Player \in \Players} \Actions{\Player}$ is the set of joint actions,. We use $\Actions{}(\WorldState) \subseteq \Actions{}$ to denote the set of joint legal actions in world state $\WorldState$. $\Transitions : \WorldStates \times \Actions{} \to \WorldStates$ is the transition function and $\Rewards{} : \WorldStates \times \Actions{} \to \mathbb{R}$ is the reward function, which corresponds to the reward of player 1. We use $\Rewards{1}(\WorldState, \Action{}) = \Rewards{}(\WorldState, \Action{})$ and $\Rewards{2}(\WorldState, \Action{}) = -\Rewards{}(\WorldState, \Action{})$ as rewards for player 1 and 2 respectively. $\Observations{}: \WorldState \times \Actions{} \times \WorldStates \to \ObservationSet{}$ is the observation function. $\ObservationSet{} = \ObservationSet{0} \times \ObservationSet{1} \times \ObservationSet{2}$ is the set of joint public and private observations. $\Observations{}$ can be factored as $\Observations{} = (\PublicObservations, \Observations{1}, \Observations{2})$, where $\PublicObservations$ is a public part of observation and $\Observations{1}, \Observations{2}$ are private parts of observations for each player. Even though we focus on simultaneous-move games, this does not limit the generality, since any sequential-move game can be converted into a simultaneous-move game by adding fictitious moves for the non-acting player in each decision node.

History $\History = \WorldState^0, \Action{}^0 \dots \Action{}^{\TrajectoryLength - 1}\WorldState^{\TrajectoryLength} \in (\WorldStates \Actions{})^* \WorldStates$ is a finite sequence of world states and actions, which starts in the initial world state $\WorldState^0 = \InitWorldState$ and for each timestep $\TimeStep \in \{0, \cdots, \TrajectoryLength - 1\}$ holds $\Action{}^\TimeStep \in \Actions{}(\WorldState^\TimeStep)$ and $\WorldState^{\TimeStep + 1} = \Transitions(\WorldState^{\TimeStep}, \Action{}^{\TimeStep})$. $\Histories$ is the set of all possible histories within the game. We use $\History \HistoryExtend \History'$  to denote that $\History'$ contains $\History$ as a prefix. We will use $\InitHistory = \InitWorldState$ to denote the initial history. Each history $\History$ ends with some world state $\WorldState^{\TrajectoryLength}$. We will sometimes use history $\History$ in the game functions instead of the world state $\WorldState^{\TrajectoryLength}$. For example $\Actions{}(\History) := \Actions{}(\WorldState^{\TrajectoryLength})$ corresponds to the set of joint legal actions in the final world state of the history. Each player $\Player$ does not observe the whole world state at each timestep, but only observes public observations $\PublicObservations$ and its private observations $\Observations{\Player}$. This means that the player may not be able to distinguish between several different histories. We will use $\Infoset{\Player} \in \Infosets{\Player}$ to denote the set of all histories consistent with the observations of the player $\Player$, which we will call the information set. $\Infosets{\Player}$ is the set of all the information sets of the player $\Player$. $\Infoset{\Player}(\History)$ is the information set that corresponds to history $\History$. Similarly, the public state $\PublicState \in \PublicStates$ is an information of an external player, which does not have private observations, so it contains all the histories consistent with public observations. Each public state contains one or more information sets for each player. 

Strategy of player $\Player$ is a function $\Strategy{\Player}: \Infosets{\Player} \to \Simplex \Actions{\Player}$ that maps each information set to a probability distribution over the actions. We will sometimes use $\Strategy{\Player}(\Infoset{\Player}, \Action{\Player})$ as a probability that player will play $\Action{\Player}$ in information set $\Infoset{\Player}$ if it follows the strategy $\Strategy{\Player}$. $\Strategy{} = (\Strategy{1}, \Strategy{2})$ is a joint strategy profile of both players. If $\History \HistoryExtend \History'$, then the reach probability of reaching history $\History'$ from history $\History$ under strategy profile $\Strategy{}$ is $\Reach{\Strategy{}}(\History' | \History) = \prod_{\History'' \Action{} \WorldState \HistoryExtend \History'} \prod_{\Player \in \Players} \Strategy{\Player}(\Infoset{\Player}(\History''), \Action{\Player})$. We also use $\Reach{\Strategy{}}(\History) := \Reach{\Strategy{}}(\History | \InitHistory)$. Any reach probability can be factored into the contribution by each player $\Reach{\Strategy{}}(\History | \History') = \prod_{\Player \in \Players} \Reach{\Strategy{}}_\Player(\History | \History')$.  Expected utility of a history $\History$ if all players follow strategy profile $\Strategy{}$ is $\Utility{\Player}^{\Strategy{}}(\History) = \sum_{\History \HistoryExtend \History' \Action{}} \Reach{\Strategy{}}(\History' | \History) \Rewards{\Player}(\History', \Action{}) \prod_{\Player \in \Players} \Strategy{\Player}(\Infoset{\Player}(\History'), \Action{\Player})$.

Best response against a strategy $\Strategy{\Player}$ is a strategy $\BRStrategy{-\Player} \in \BRSet{-\Player}$, which maximizes the opponent's utility $\Utility{-\Player}^{(\Strategy{\Player}, \BRStrategy{-\Player})}(\InitHistory) \geq \Utility{-\Player}^{(\Strategy{\Player}, \Strategy{-\Player}')}(\InitHistory)$ for any $\Strategy{-\Player}'$. We use a $-\Player$ here to symbolize the other player than $\Player$, which is a standard notation in games. If all players play a best response to each other, the resulting strategy profile is known as Nash Equilibrium $\Nash{}$ \cite{nash1950, gtessentials1952}. In two-player zero-sum games, this is usually the sought after solution concept. As a metric to evaluate quality of a strategy, we use exploitability $\Exploitability(\Strategy{\Player}) = \Utility{-\Player}^{(\Strategy{\Player}, \BRStrategy{-\Player})}(\InitHistory) - \Utility{-\Player}^{\Nash{}}(\InitHistory)$, which is how much can opponent gain, when it plays best response as compared to the Nash equilibrium. In two-player zero-sum games, the exploitability is always nonnegative and is zero if and only if the $\Strategy{\Player}$ is a part of Nash equilibrium.

\section{Learning the Game Model}
\label{sec:modelfree}
In perfect information games, players possess complete knowledge of the current game state represented by a history $\History$. Consequently, search algorithms initiate from a single, known root state, simplifying the search. In contrast, imperfect information games (IIGs) grant players only partial observability through an information set $\Infoset{\Player}$, which typically corresponds to multiple possible underlying world states. As established in prior works, approximating optimal strategies via look-ahead reasoning in IIGs requires a more sophisticated approach than in perfect information settings \cite{valuefunctions2023, deepstack2017}.

Specifically, it is insufficient to restrict the reasoning to only those histories consistent with the player's $\Player$ information set. Instead, the reasoning must encompass all histories that share the public state $\PublicState$. This necessity arises because sound reasoning algorithms compute strategies for all players simultaneously, aiming for mutual best responses characteristic of an equilibrium. Consider this situation in two-player Poker: if player $\Player$ holds two Kings, their information set $\Infoset{\Player}$ includes all histories consistent with this hand but with varying opponent hands. A reasoning restricted only to those histories would implicitly grant the opponent knowledge of $\Player$'s hand when computing opponent's strategy, leading to suboptimal strategies. Therefore, the look-ahead reasoning must operate over the broader set of histories consistent with public state to compute valid equilibrium strategies \cite{deepstack2017, studentofgames2023, valuefunctions2023,milec2024}.

We adopt a reinforcement learning paradigm where an agent learns from interaction with the environment. During training, we assume access to a game simulator capable of generating the whole game trajectories. During testing (gameplay), the agent receives only its own information set $\Infoset{\Player}$ at each step and must rely entirely on its learned model to plan, without access to the simulator or explicit rules. This means that the agent does not use any domain-specific knowledge. at any point.

We propose a model inspired by MuZero \cite{muzero2020} but adapted for the IIG setting, which requires additional structures necessary to model the imperfect information. Our model comprises three core learnable functions, parameterized by $\NeuralParameters$:

\begin{itemize}
  \item Representation function $\RepresentationFunction_\NeuralParameters: \Infosets{\Player} \to \Abstracted{\Infosets{\Player}}$. Maps a player $\Player$'s potentially high-dimensional information set $\Infoset{\Player}$ to a fixed-size latent representation $\Abstracted{\Infoset{\Player}} \in \Abstracted{\Infosets{\Player}}$. 
  \item Dynamics function $\DynamicsFunction_\NeuralParameters: \Abstracted{\Infosets{1}} \times \Abstracted{\Infosets{2}} \times \Actions{1} \times \Actions{2} \to \Abstracted{\Infosets{1}} \times \Abstracted{\Infosets{2}} \times \RealNumbers \times \BinaryNumbers$. Given the latent representations for all players ($\Abstracted{\Infoset{1}}, \Abstracted{\Infoset{2}}$) and the joint action taken ($\Action{1}, \Action{2}$), this function predicts the resulting next latent representations for both players, the immediate reward $\Reward{}$ (e.g., for player 1), and a binary termination flag $\TerminalLogit$. This models the joint evolution of the game across possible hidden states. We use $\BinaryNumbers = \{0, 1\}$.
  \item Legal actions function $\LegalActionsFunction_\NeuralParameters: \Abstracted{\Infosets{\Player}} \to \BinaryNumbers^{|\Actions{\Player}|}$. Predicts the mask of legal actions $\Actions{\Player}$ available to player $\Player$ from their latent representation $\Abstracted{\Infoset{\Player}}$. This is crucial for constraining the look-ahead reasoning only to the feasible parts of the game.
\end{itemize}

Each training episode a game trajectory is sampled. This trajectory is $\History = \InitWorldState \Action{}^0 \dots \Action{}^{\TrajectoryLength - 1}\WorldState^{\TrajectoryLength}$, where $\Action{}^{\TimeStep}= (\Action{1}^{\TimeStep}, \Action{2}^{\TimeStep})$ is the joint action at step $\TimeStep$. For any sub-history $\History^\TimeStep \HistoryExtend \History$, the simulator provides the true information sets $\Infoset{\Player}^\TimeStep(\History^\TimeStep)$, legal actions $\Actions{\Player}^\TimeStep(\Infoset{\Player}^\TimeStep)$, and the reward $\Reward{}^\TimeStep$. The model is trained to predict these quantities through recurrent application of its components.

Starting from an initial latent state $\Abstracted{\Infoset{\Player}^{\TimeStep, 0}} = \RepresentationFunction_\NeuralParameters(\Infoset{\Player}^\TimeStep)$, the dynamics function is unrolled for $\UnrollStep$ steps using the actual actions from the trajectory:

$$\Abstracted{\Infoset{1}^{\TimeStep, \UnrollStep + 1}}, \Abstracted{\Infoset{2}^{\TimeStep, \UnrollStep + 1}}, \Abstracted{\Reward{}^{\TimeStep, \UnrollStep + 1}}, \Abstracted{\TerminalLogit{}^{\TimeStep, \UnrollStep + 1}} = \DynamicsFunction_\NeuralParameters(\Abstracted{\Infoset{1}^{\TimeStep, \UnrollStep}}, \Abstracted{\Infoset{2}^{\TimeStep, \UnrollStep}}, \Action{1}^{\TimeStep + \UnrollStep}, \Action{2}^{\TimeStep + \UnrollStep})$$

Here, $\Abstracted{\Infoset{\Player}^{\TimeStep, \UnrollStep}}$ is the predicted latent state after $\UnrollStep$ unrolls from step $\TimeStep$, and $\Abstracted{\Reward{}^{\TimeStep, \UnrollStep}} $ and $\Abstracted{\TerminalLogit{}^{\TimeStep, \UnrollStep}}$ are the predicted reward and termination logit. The legal actions function predicts logits $\Abstracted{\ActionLogit{\Player}^{\TimeStep}} = \LegalActionsFunction_\NeuralParameters(\Abstracted{\Infoset{\Player}^{\TimeStep, 0}})$ from the initial latent state.

The model parameters $\NeuralParameters$ are optimized by minimizing a combined loss function over the trajectory:
\begin{align}
\begin{split}
\ModelLoss = \sum_{\TimeStep = 0}^{\TrajectoryLength - 1} \bigg[ &\sum_{\Player \in \Players} \underbrace{\LegalActionsLoss(\Abstracted{\ActionLogit{\Player}^{\TimeStep}}, \Actions{\Player}^{\TimeStep})}_\text{Legal Action Prediction} \\
+& \sum_{\UnrollStep = 1}^{\TrajectoryLength - \TimeStep} \Big( \underbrace{\TerminalLoss(\Abstracted{\TerminalLogit^{\TimeStep, \UnrollStep}}, \mathbb{I}[t+k = \TrajectoryLength])}_\text{Termination Prediction} + \underbrace{\RewardLoss(\Abstracted{\Reward{}^{\TimeStep, \UnrollStep}}, \Reward{}^{\TimeStep + \UnrollStep})}_\text{Reward Prediction} + \sum_{\Player \in \Players} \underbrace{\DynamicsLoss(\Abstracted{\Infoset{\Player}^{\TimeStep, \UnrollStep}}, \RepresentationFunction_\NeuralParameters(\Infoset{\Player}^{\TimeStep + \UnrollStep})) }_\text{Latent State Prediction} \Big) \bigg]
\end{split}
\end{align} 
where $\LegalActionsLoss$ and $\TerminalLoss$ are binary cross-entropy losses, while $\RewardLoss$ and $\DynamicsLoss$ are mean squared errors. The target for the dynamics loss is the latent representation of the actual subsequent information set. 

Minimizing $\ModelLoss$ trains the functions $\RepresentationFunction_\NeuralParameters, \DynamicsFunction_\NeuralParameters, \LegalActionsFunction_\NeuralParameters$ to collectively act as a learned simulator. This learned model captures the necessary dynamics within the game, enabling test-time look-ahead reasoning algorithms (discussed in \cref{sec:depthlimited}) to effectively plan over the required set of public states without recourse to the original game rules or simulator.
\section{Learning the Abstract Model}
\label{sec:abstraction}
A primary limitation of sound look-ahead reasoning in imperfect information games is the potential size of the public states, as the number of information sets consistent with public information may be exponential in the history length, making the look-ahead reasoning intractable \cite{deepstack2017,studentofgames2023}.

Although traditional abstraction techniques often rely on domain expertise or require offline enumeration and analysis of all information sets \cite{cermak2020,kroer2018,ganzfried2014,brown2015,bard2014,johanson2013}, we aim to learn domain-independent abstraction directly from the game experience during training. Our goal is to partition the vast space of information sets sharing a public state into a manageable number $\AbstractionLimit$ of abstract information sets, enabling tractable reasoning. Such an abstraction may contain imperfect recall as discussed in \cref{sec:limitations}.

Consider Texas Hold'em Poker as an example. A player might hold any of the $1326$ private hands, each corresponding to a different information set. Our abstraction aims to represent this multitude using only $\AbstractionLimit$ representatives, learned based on similarity within the training process rather than predefined rules.

We adapt the model from \cref{sec:modelfree}, but we will use mechanisms inspired by online clustering to limit the amount of information sets in each public state. We hypothesize that information sets behaving similarly, e.g. having similar optimal strategies or leading to similar future states, should be grouped. We formalize this using a function $\PropertiesFunction: \Infosets{\Player} \to \RealNumbers^K$, which maps any information set to a $K$-dimensional space, in which the clustering will be performed.

\begin{figure} 
    \centering
    \includegraphics[width=0.5\linewidth]{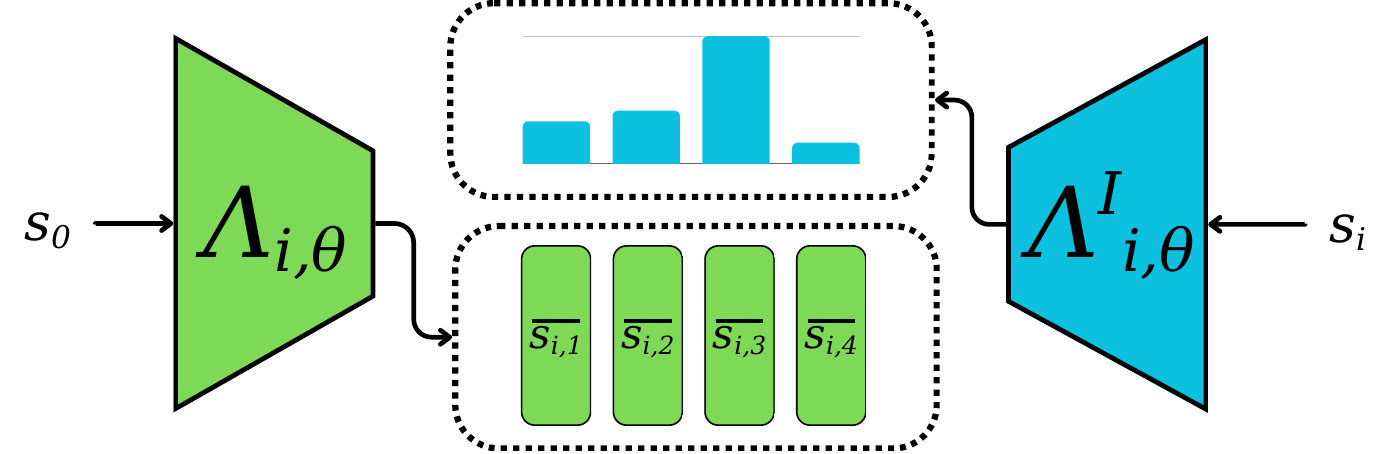}
    \caption{The public state and information set representations functions of player $\Player$. First the $\AbstractionFunction_{\Player, \NeuralParameters}$ predicts 4 abstract information sets and then $\RepresentationFunction_{\Player, \NeuralParameters}$ predicts the probability distribution over those abstractions.}
    \label{fig:abstraction_network}
\end{figure}

To satisfy the condition that each public state consists of at most $\AbstractionLimit$ information sets of each player, we split the representation function into two parts as shown in \cref{fig:abstraction_network}. The public state representation $\AbstractionFunction_{\Player, \NeuralParameters}: \PublicStates \to \Abstracted{\Infosets{\Player}}^{\AbstractionLimit}$  maps a public state $\PublicState$ to $\AbstractionLimit$ latent abstract information sets for a player $\Player$. The information set representation $\RepresentationFunction_{\Player, \NeuralParameters}: \Infosets{\Player} \to \Simplex\Abstracted{\Infosets{\Player}}$ maps a real information set to the probability distributions on the abstract information sets provided by $\AbstractionFunction_{\Player, \NeuralParameters}$. Crucially, despite $\RepresentationFunction_{\Player, \NeuralParameters}$ providing a probability distribution, we enforce the many-to-one mapping, so we represent any real information set by the single abstract information set, corresponding to the highest probability. This representative will then be used for training dynamics $\DynamicsFunction_\NeuralParameters$ and legal actions $\LegalActionsFunction_\NeuralParameters$. We opted to this, so that the dynamics function constructs the search tree, which is compatible with look-ahead reasoning algorithms like Counterfactual Regret Minimization.
 
In order to perform the clustering, we require the same function $\PropertiesFunction$ as for the real information sets. We introduce $\PropertiesFunction_{\NeuralParameters}: \Abstracted{\Infosets{\Player}} \to \RealNumbers^{K}$, which will be trained to predict this clustering property for each abstract information set. The dynamics function $\DynamicsFunction_{\NeuralParameters}$ and legal actions function $\LegalActionsFunction_{\Player, \NeuralParameters}$ are defined as in \cref{sec:modelfree}.

To learn the abstraction that approximates the proposed clustering based on $\PropertiesFunction$, we train $\AbstractionFunction_{\Player, \NeuralParameters}, \RepresentationFunction_{\Player, \NeuralParameters}$ and $\PropertiesFunction_\NeuralParameters$ jointly. Each training step, the simulator provides trajectory $\History = \InitWorldState \Action{}^0 \dots \Action{}^{\TrajectoryLength - 1}\WorldState^{\TrajectoryLength}$. $\PublicState^{\TimeStep}$ $\Infoset{\Player}^\TimeStep$ represents the public state and information set of player $\Player$ at time $\TimeStep$ in the same trajectory.

The training involves two additional losses, the first one $\AbstractionLoss$ updates the $\AbstractionFunction_{\Player, \NeuralParameters}$ and $\PropertiesFunction_\NeuralParameters$ using a soft clustering objective similar to fuzzy c-means \cite{bezdek1984fcm}. It minimizes a mean squared error between real and abstract properties weighted by the softmax. The second loss $\InfosetModelLoss$ updates the $\RepresentationFunction_\NeuralParameters$ by minimizing the cross entropy loss between the predicted probability distribution over the abstract information sets and the one-hot encoded vector of the abstract information set that is the nearest neighbor of the real information set based on the $\PropertiesFunction$.
\begin{equation}
\AbstractionLoss = \sum_{\TimeStep}^{\TrajectoryLength - 1} \sum_{\Player \in \Players} \sum_{\Abstracted{\Infoset{\Player}^\TimeStep} \in \AbstractionFunction_{\Player, \NeuralParameters}(\PublicState^{\TimeStep})} ||\PropertiesFunction_{\NeuralParameters}(\Abstracted{\Infoset{\Player}^\TimeStep}) - \PropertiesFunction(\Infoset{\Player}^\TimeStep)||^2 \frac{e^{ - \SoftmaxTemperature ||\PropertiesFunction_{\NeuralParameters}(\Abstracted{\Infoset{\Player}^\TimeStep}) - \PropertiesFunction(\Infoset{\Player}^\TimeStep)||^2}}{\sum_{\Abstracted{\Infoset{\Player}^\TimeStep}' \in \AbstractionFunction_{\Player, \NeuralParameters}(\PublicState^{\TimeStep})} e^{ - \SoftmaxTemperature ||\PropertiesFunction_{\NeuralParameters}(\Abstracted{\Infoset{\Player}^\TimeStep}') - \PropertiesFunction(\Infoset{\Player}^\TimeStep)||^2}}
\end{equation} 
\begin{equation}
\InfosetModelLoss = \sum_{\TimeStep}^{\TrajectoryLength - 1} \sum_{\Player \in \Players} \sum_{\Abstracted{\Infoset{\Player}^\TimeStep} \in \AbstractionFunction_{\Player, \NeuralParameters}(\PublicState^{\TimeStep})}  \InfosetLoss(\Abstracted{\Infoset{\Player}^\TimeStep}, \Infoset{\Player}^\TimeStep)
\end{equation}
The $\SoftmaxTemperature$ controls the softness of the clustering and as $\SoftmaxTemperature \to \infty$ the clustering becomes hard. The gradients from $\AbstractionLoss$ propagate through both $\PropertiesFunction_\NeuralParameters$ and $\AbstractionFunction_\NeuralParameters$, but the gradients from $\InfosetModelLoss$ are only propagated through $\RepresentationFunction_\NeuralParameters$. 
\begin{equation}
\ModelAbstractionLoss = \ModelLoss + \AbstractionLoss + \InfosetModelLoss
\end{equation}
The overall loss also includes the model learning loss from \cref{sec:modelfree}. Importantly, $\ModelLoss$ is computed using the dynamics based on the selected abstract information set. Furthermore, the gradients from $\ModelLoss$ are not backpropagated through $\AbstractionFunction_\NeuralParameters$, $\RepresentationFunction_\NeuralParameters$ or $\PropertiesFunction_\NeuralParameters$. This decouples the learning of the abstraction structure from the learning of the model dynamics.

\section{Depth-Limited Solving}
\label{sec:depthlimited}
While the learned model and abstraction allow reconstructing the whole game tree of the abstracted game, practical applications in large imperfect information games rely on depth-limited reasoning combined with a learned value function to estimate payoffs beyond the reasoning horizon. Defining and training optimal value function for imperfect information games is challenging, as they theoretically depend on belief states \cite{valuefunctions2023,studentofgames2023}. The belief states are public states with corresponding probability distribution of reaching each history. Such value function is often trained by repeatedly sampling varied belief states and performing depth-limited reasoning in each of them.  Such value function may be trained even in our model, but it would require two-phase training process, where first phase trains the abstraction and second the value function.

To enable single-phase training we integrate an approximate value function based on the multi-valued states approach \cite{mvs2018,sepot2024,milec2025}, which results in an algorithm we call \textit{Learned Abstract Model for Imperfect-information Reasoning} (\AlgorithmName{}), which consists of the parts described in Sections \ref{sec:modelfree} and \ref{sec:abstraction} and these additional components:

\begin{itemize}
  \item Strategy function $\StrategyFunction_{\NeuralParameters} : \Infosets{\Player} \to \Simplex\Actions{}$, which for given information set from the original game returns the strategy trained with some policy-gradient algorithm, like RNaD \cite{stratego2022}.
  \item Transformations function $\TransformationsFunction_{\NeuralParameters} : \Infosets{\Player} \to \RealNumbers^{|\Actions{\Player}| \times |\Transformations|}$, representing $\Transformations$ characteristic directions in strategy space explored by the policy-gradient algorithm during training. For a single transformation $\Transformation \in \TransformationsFunction_{\NeuralParameters}$ the transformed strategy is computed locally as $\Strategy{\Player}^\Transformation(\Infoset{\Player}, \Action{\Player}) = \Strategy{\Player}(\Infoset{\Player}, \Action{\Player}) + \Transformation(\Infoset{\Player}, \Action{\Player})$. The resulting strategy $\Strategy{\Player}^\Transformation(\Infoset{\Player})$ is then normalized \cite{sepot2024}.
  \item Value function $\ValueFunction_{\NeuralParameters} : \Abstracted{\Infosets{\Player}} \to \RealNumbers^{|\Transformations| \times |\Transformations|}$, which approximates the expected value of each combination of transformed strategies between players \cite{sepot2024}.
\end{itemize}

We train $\StrategyFunction_{\NeuralParameters}$ and $\TransformationsFunction_{\NeuralParameters}$ using real information sets from sampled trajectories. However, $\ValueFunction_{\NeuralParameters}$ is trained using the corresponding joint abstract information sets. Multiple real histories can map to the same abstract state with varying reach probabilities under the sampling strategy. This introduces bias if the value function aims to represent the expected value under the target strategy. While importance sampling could correct this, we found in our experiments it did not affect the results in any significant way. This is most likely due to transformations being just a heuristic approach to approximate different parts of the strategy space.

We used Regularized Nash Dynamics (RNaD) as the policy-gradient algorithm, which includes Neural Replicator Dynamics (NeuRD) loss for strategy training and mean squared error for the associated baseline value function $\PolicyGradientLoss$ \cite{neurd2020,poincare2021,stratego2022}. Note, that the value function from the RNaD cannot be used as a value function for look-ahead reasoning, because it does not work with the belief states. We use mean squared error for the training of the value function $\ValueFunctionLoss$ and the targets are computed by the V-trace \cite{impala2018,sepot2024}, which estimates the value of different policies in an off-policy setting.

Following \cite{sepot2024}, transformations represent the strategy changes during training. For each player $\Player$ we compute the difference vector $\StrategyDirection_{\Player}^{\TimeStep} = \Strategy{\Player}^{\TimeStep, \text{NEW}} - \Strategy{\Player}^{\TimeStep, \text{OLD}}$, where $\Strategy{\Player}^{\TimeStep, \text{OLD}}$ and $\Strategy{\Player}^{\TimeStep, \text{NEW}}$ are concatenated strategies along the whole trajectory before and after the training step. Instead of the hard clustering proposed originally, we use the soft clustering from \cref{sec:abstraction}.

\begin{equation}
\TransformationsLoss = \sum_{\Player \in \Players} \sum_{\Transformation_\Player \in \Transformations} ||\Transformation_\Player - \StrategyDirection_{\Player}^{\TimeStep}||^2 \frac{e^{ - \SoftmaxTemperature ||\Transformation_\Player - \StrategyDirection_{\Player}^{\TimeStep}||^2}}{\sum_{\Transformation_\Player' \in \Transformations} e^{ - \SoftmaxTemperature ||\Transformation_\Player' - \StrategyDirection_{\Player}^{\TimeStep}||^2}}
\end{equation}
Since transformation loss depend on strategy changes after the step of policy-gradient training, we use a two-step update per episode. First we update the Policy-gradient algorithm and then all the other components of the model.
\begin{equation}
\Loss_\NeuralParameters^1 = \PolicyGradientLoss 
\end{equation}
\begin{equation}
\Loss_\NeuralParameters^2 = \ModelAbstractionLoss +\ValueFunctionLoss + \TransformationsLoss
\end{equation}

At test time, \AlgorithmName{} employs continual resolving \cite{deepstack2017,studentofgames2023} for the acting player $\Player$. It starts in the initial public state with a single joint abstract information set. The following process then repeats until the terminal state is reached. The algorithm constructs the depth-limited game tree using the trained dynamics and legal actions functions $\DynamicsFunction_\NeuralParameters, \LegalActionsFunction_\NeuralParameters$. At the depth-limit in non-terminal states it adds additional decision layer with $\Transformations$ actions for both players. Each combination of actions corresponds to joint transformations and it leads to a terminal state with rewards corresponding to $\ValueFunction_\NeuralParameters$. In this tree, the algorithm performs look-ahead reasoning with Counterfactual Regret Minimization+ (CFR+) \cite{cfr2007,CFRPlus2014}. This results in a strategy in each abstract information set in current public state $\PublicState$. The real current information set $\Infoset{\Player}$ is mapped to the abstracted one $\Abstracted{\Infoset{\Player}}$ with $\AbstractionFunction_{\Player, \NeuralParameters}$ and $\RepresentationFunction_{\Player, \NeuralParameters}$. The CFR strategy in $\Abstracted{\Infoset{\Player}}$ is then used to sample the action and move into a new public state $\PublicState$ and ${\Infoset{\Player}}$. Out of the previous game tree, all the abstracted information sets in $\PublicState$ are used to create a new gadget game \cite{cfrd2014}. Both counterfactual values and the reaches of the resolving player $\Player$ are reused from the previous subtree. The algorithm repeats this process until it reaches the terminal state.
\section{Experiments}
\label{sec:experiments}
\subsection{Exploitability in Smaller Games}
\label{sec:exploitability_experiment}
To ensure that the strategies found by \AlgorithmName{} approximate Nash equilibria, we applied it in games small enough to compute exact exploitability, which serves as a distance from the Nash equilibrium in two-player zero-sum games. For various abstraction sizes $\AbstractionLimit$ and different properties for clustering $\PropertiesFunction$ we trained \AlgorithmName{} with 10 different random seeds for 100,000 episodes. In all of our experiments we used Regularized Nash Dynamics (RNaD) to train the strategy for multi-valued states value function. The rules of the games used in experiments are in Appendix~\ref{sec:app_game_rules}.

\begin{figure}[htbp]
    \centering
        \begin{subfigure}[b]{0.49\textwidth}
            \centering
            \includegraphics[width=\textwidth]{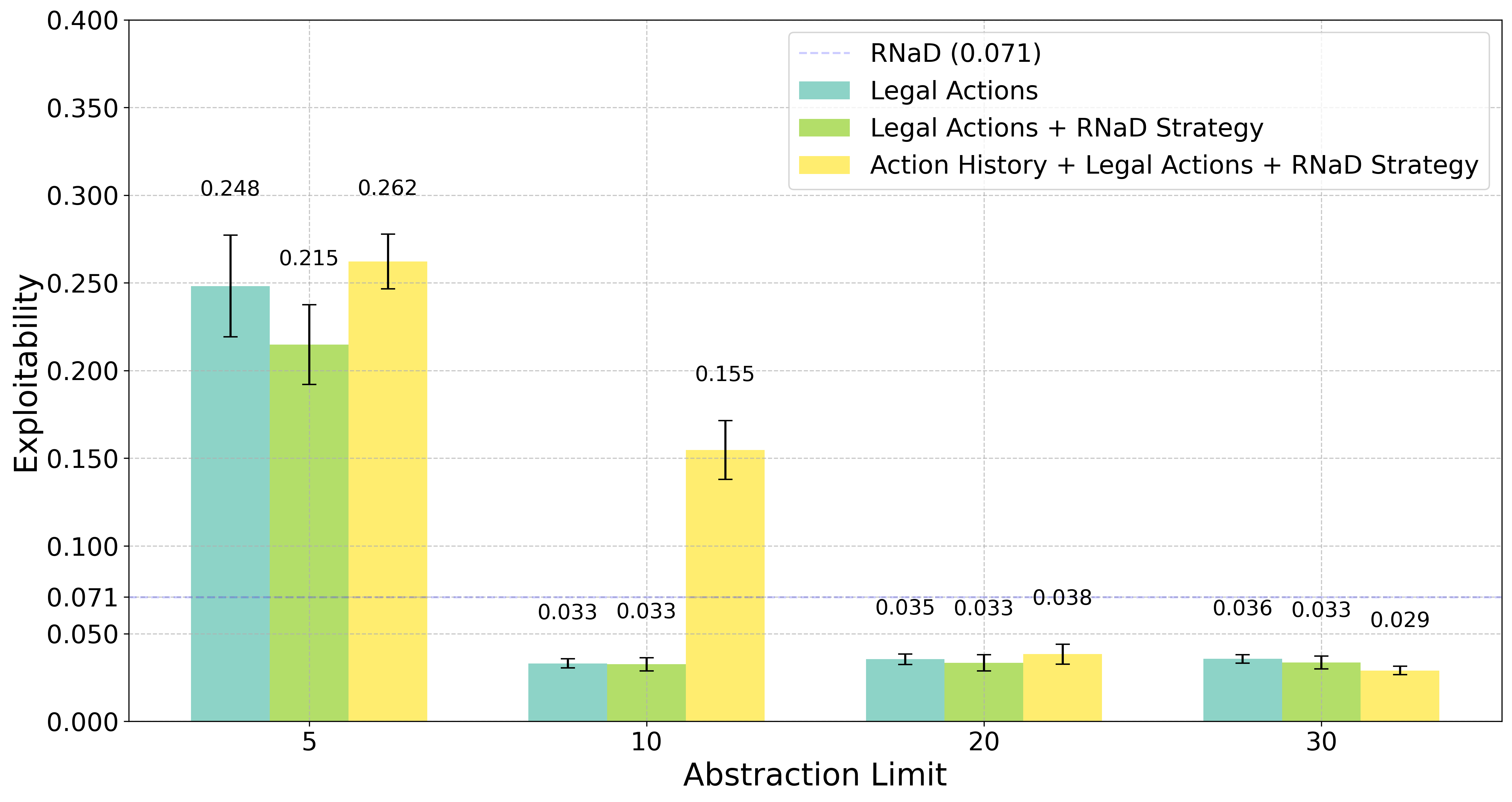}
            \caption{Imperfect Information Goofspiel 5}
            \label{fig:goofspiel_5_subgame}
        \end{subfigure}%
        \hfill
        \begin{subfigure}[b]{0.49\textwidth}
            \centering
            \includegraphics[width=\textwidth]{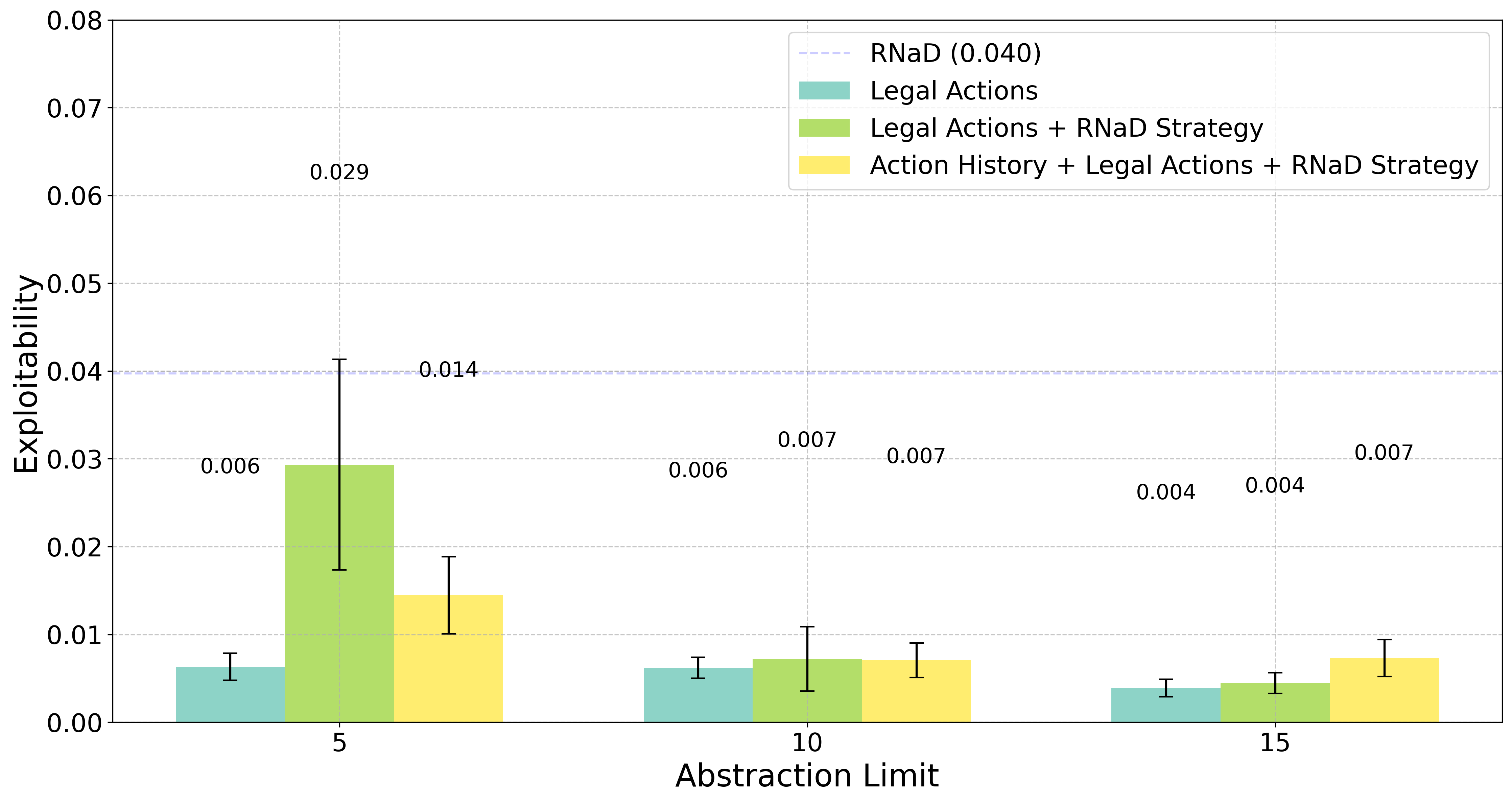}
            \caption{Imperfect Information Oshi-Zumo 3,5}
            \label{fig:oshi_zumo_3_5_subgame}
        \end{subfigure}
    \caption{Exploitability of \AlgorithmName{} in a different games by using continual resolving with depth-limit 1 in each subgame with different choice of abstraction limit $\AbstractionLimit$ and $\PropertiesFunction$. The largest public state in II Goofspiel 5 contains 30 infosets and in Oshi-Zumo 3,5 it contains 625 information sets.}
    \label{fig:exploitability}
\end{figure}

Starting at episode 80,000 we have computed exploitability every 1000 episodes. In each public state, the algorithm uses the trained functions to construct the depth-limited subgame with depth limit 1. Then this subgame is solved using CFR+ and the strategy from abstract infosets is mapped to real ones. Then we compute the exploitability of the final composed strategy in the original game. The results for different $\PropertiesFunction$ and $\AbstractionLimit$ are displayed in \cref{fig:exploitability}. 

We used three different $\PropertiesFunction$, which served as a basis for clustering, first the legal actions, second the legal actions with the current RNaD strategy, and third the legal actions, RNaD strategy, and the player's action history. Each choice of $\PropertiesFunction$ is capable of outperforming the concurrently trained RNaD with the same amount of training episodes given sufficient $\AbstractionLimit$. When using the action history as $\PropertiesFunction$, each information set is uniquely defined in a given public state, which means that, with sufficient capacity, the dynamics network should model the underlying game. This was evaluated in II Goofspiel 5 with $\AbstractionLimit=30$, which indeed has mapping one-to-one for each abstract and information set. The main reason why the exploitability is greater than 0 arises from the discrepancies in the rewards produced by the dynamics function.

\subsection{Head-to-head in Large Games}
\label{sec:heads_experiment}
The main usage of \AlgorithmName{} is in very large games, where the exact exploitability cannot be computed. This experiment evaluates \AlgorithmName{} in this exact setting, where we compare it with Regularized Nash Dynamics (RNaD) in head-to-head play. For each game and $\PropertiesFunction$, we have trained \AlgorithmName{} with 3 different random seeds for 3 million episodes. Similarly, we have trained RNaD with 6 different random seeds for the same number of episodes. The hyperparameter settings remained the same for each game. Then we matched each trained seed of \AlgorithmName{} with each trained seed of RNaD and played more than 100,000 matches. Note that when we use $\PropertiesFunction =$ RNaD strategy, it is the strategy that is learned concurrently for the value function and not the RNaD strategy against which the algorithm is compared later. The resulting win rates with 2-sigma error bars are in \cref{tab:heads_to_heads}.

\begin{table}[t]
\centering
\begin{tabular}{|l|c|c|c|}
\hline
\textbf{Algorithm} & \textbf{II Goofspiel 10} & \textbf{II Goofspiel 13} & \textbf{II Goofspiel 15} \\
\hline
\AlgorithmName{} $\PropertiesFunction=$ Legal actions & 54.47 $\pm$ 0.25 \% & 60.68 $\pm$ 0.34 \% & 80.49 $\pm$ 0.26 \% \\
\hline
\AlgorithmName{} $\PropertiesFunction=$ RNaD strategy & 61.60 $\pm$ 0.29 \% & 58.33 $\pm$ 0.27 \% & 61.80 $\pm$ 0.36 \% \\
\hline
\end{tabular}
\caption{Average win rate with 2-sigma error bars of \AlgorithmName{} against RNaD in different games.}
\label{tab:heads_to_heads}
\end{table}

\AlgorithmName{} consistently outperforms RNaD in each of the tested games. Imperfect Information Goofspiel is known for its large public states, so continual resolving without abstractions is not applicable. SePoT \cite{sepot2024} was also evaluated in such large domains, but we did not compare against it, as it only uses CFR if the subgame is small enough. The authors showed that in II Goofspiel 13, SePoT has a win rate of only 52\%, which is likely caused by not resolving almost any subgame due to the limit on the size of the subgame. Furthermore, we show that even in larger games than those tested with SePoT, \AlgorithmName{} improves over RNaD even more.

\section{Limitations and Future Work}
\label{sec:limitations}
\AlgorithmName{} advances the scalability of the continual resolving paradigm to larger games by integrating learned models and abstractions. However, it presents several interesting avenues for future research.

The computational complexity of the look-ahead reasoning, when using CFR is linear in the amount of information sets in the game. This complexity still remains, as \AlgorithmName{} only reduces the size of the game in each public state but uses CFR in the abstract game. Each subgame \AlgorithmName{} construct with depth $\DepthLimit$ contains at most $\sum_{\DepthIterator \in \{0, \dots, D\}} \AbstractionLimit^2 |\Actions{}|^{2\DepthIterator}$ unique nodes and at most $\AbstractionLimit^2 |\Actions{}|^{2\DepthLimit} \Transformations^2$ terminal histories, where $\Transformations$ is the amount of transformations. Most of the subgames would be smaller, but this is the main limitation of \AlgorithmName{} as it limits how large $\AbstractionLimit$ can be.

Currently, \AlgorithmName{}'s dynamics network $\DynamicsFunction_\NeuralParameters$ does not explicitly model chance nodes within the game. \AlgorithmName{} could still be applied in games with chance events, but this is not an intended setting as the absence of chance nodes will result in poor abstraction, regardless of the abstraction capacity $\AbstractionLimit$. Although algorithms like Stochastic MuZero \cite{stochasticmuzero2022} demonstrate that modeling chance is feasible in learned models for perfect information games, integrating chance nodes into our framework, particularly in conjunction with learned abstractions, requires careful consideration and is a key area for future work.

The effectiveness of the learned abstraction depends on the chosen property function $\PropertiesFunction$ for clustering. In games without chance, if this property function could perfectly distinguish two different information sets, $\AbstractionLimit$ is greater than the size of the largest public state and neural networks have sufficient capacity, then \AlgorithmName{} learns a near-perfect model of the game. When $\AbstractionLimit$ is reduced to achieve scalability, the learned abstraction may not capture all crucial strategic nuances, which can affect the strength of the derived strategy. In experiments, we have used simple proxies present in any game and it still yielded strong performance. Ideally, $\PropertiesFunction$ would also be learned during the training process.

Reducing $\AbstractionLimit$ necessarily introduces imperfect recall, meaning that the player in the abstract game may "forget" information it previously knew. Algorithms like Counterfactual Regret Minimization (CFR) guarantee convergence to a Nash equilibrium strategy only in perfect recall games. While CFR's convergence is not generally guaranteed in imperfect recall settings, it has been shown for subclasses such as A-loss recall games \cite{cermak2020}. Our abstractions are A-loss recall games if the public observations in the original game depend only on prior public information and the actions taken in the current round. Games like Imperfect Information Goofspiel or Oshi-Zumo satisfy these conditions, but many others, including Battleships, Dark Chess, or Stratego, do not. Thus, for games that do not fall into the A-loss recall category after abstraction, theoretical convergence guarantees for CFR-based methods within \AlgorithmName{} are not assured.

\AlgorithmName{} focuses on abstracting information sets but does not inherently abstract action spaces. In games with very large or continuous action spaces (e.g., bet sizing in Poker), the sheer number of actions can remain a bottleneck for the look-ahead reasoning, regardless of the information set abstraction. While action abstractions have been extensively studied \cite{brownabstractions2014,libratus2018,actionabstraction2024}, integrating them with \AlgorithmName{} is out of the scope of single paper, but it presents another direction for future improvement.


\section{Conclusion}
\label{sec:conclusion}
In this paper, we have introduced \AlgorithmName{}, an algorithm designed to enable look-ahead reasoning by learning the model of the game with a suitable abstraction directly from experience, without the need for any domain-specific knowledge.

Our core contributions are fourfold. First, we have proposed a method for learning the fundamental components of a model of dynamics of imperfect information games without chance. Second, we developed a technique for automatically learning an abstraction by clustering information sets, effectively reducing the size of the game. Third, we integrated these components with a learning of value function in the abstracted game that enables depth-limited look-ahead reasoning. Lastly, we demonstrate how \AlgorithmName{} facilitates the continual resolving paradigm by performing a depth-limited look-ahead reasoning in each decision node encountered.

We empirically verify that, when given sufficient capacity, \AlgorithmName{} learns a nearly perfect model. Still, the game-playing capabilities degrade gracefully when the abstraction capacity is reduced. We have also shown that \AlgorithmName{} manages to perform look-ahead reasoning even in intractably large public states. Thanks to that, it achieved up to 80\% win rate in large games compared to RNaD, a strong baseline that was successfully used to create a human-expert level player in Stratego.

The primary impact of \AlgorithmName{} lies in scaling look-ahead reasoning techniques to larger games by learning abstraction and its model directly from experience. This overcomes the most notable limitations of the continual resolving paradigm, which requires each subgame to be tractable and to have explicit access to the game model. \AlgorithmName{} has several limitations, which can be addressed in future work. Still, \AlgorithmName{} is, to the best of our knowledge, the first algorithm that enables look-ahead reasoning in large-scale games like Imperfect Information Goofspiel 15 without any domain-specific knowledge, substantially outperforming the model-free policy-gradient algorithm.

\section*{Acknowledgments}
This research is supported by Czech Science Foundation
(GA25-18353S) and Grant Agency of the CTU in Prague
(SGS23/184/OHK3/3T/13). Computational resources were supplied
by (e-INFRA CZ LM2018140) supported by the Ministry of
Education, Youth and Sports of the Czech Republic and also
(CZ.02.1.01/0.0/0.0/16 019/0000765)

\bibliography{references}
\bibliographystyle{iclr2026_conference}

\appendix
\section{Experimental details}
\label{app:experimental_details}
Our implementation of \AlgorithmName{} was done in Python 3.12.3 \cite{python2009} with libraries from JAX ecosystem \cite{jax2018,optax2020,flax2020} for automatic differentiation and GPU acceleration. Our implementation of RNaD is derived out of the implementation in OpenSpiel \cite{openspiel2019}. All experiments were run in cluster with several GPUs Nvidia Tesla A100 and CPUs AMD EPYC 7543 with 1TB memory. Each experiment always used only one CPU core and one GPU at most. The training used at most 8GBs of memory per training. Gameplay evaluation and exploitability computation took at most 64GBs of memory. The hyperparameters used are in \cref{tab:hyperparameters,tab:hyperparameters_experiment}

\subsection{Soft clustering changes}
\label{app:soft_cluster}
When we have used the soft clustering as described in \cref{sec:abstraction}. However, we have encountered that in some instances when the abstraction limit $\AbstractionLimit$ was greater than the amount of information sets in a public state, each cluster center. This caused a problem with the dynamics network, which sometimes mapped to two different continuations and therefore revealed some information to the player. Since some information sets are sampled more often than others, it may happen that the soft clustering pulls several clusters more to this sample than to others. To avoid these problems, we have introduced three changes to the clustering loss. First, we added an additional loss $\RepulsionLoss$ that repels the clusters that are closer than $\RepulsionConstant$. Second, we changed $\AbstractionLoss$ so that if some cluster is closer than $\HardClusteringDist$, we changed the soft clustering to hard, e.g. we move only the closest cluster, for that particular sample. Third, we added Gaussian noise to each sampled point with mean 0 and scale $\sigma = 0.02$.

\begin{equation}
    \RepulsionLoss = \sum_{\TimeStep}^{\TrajectoryLength - 1} \sum_{\Player \in \Players} \sum_{\Abstracted{\Infoset{\Player}^\TimeStep} \in \AbstractionFunction_{\Player, \NeuralParameters}(\PublicState^{\TimeStep})} \sum_{\Abstracted{\Infoset{\Player}'^{\TimeStep }} \in \AbstractionFunction_{\Player, \NeuralParameters}(\PublicState^{\TimeStep})} \max\{0, \RepulsionConstant - ||\PropertiesFunction_{\NeuralParameters}(\Abstracted{\Infoset{\Player}^\TimeStep}) - \PropertiesFunction_{\NeuralParameters}(\Abstracted{\Infoset{\Player}'^{\TimeStep }} )||^2\}
\end{equation}

\subsection{Public state decoder}
With the abstraction loss $\AbstractionLoss$ as we defined it in \cref{sec:abstraction}, it may happen that two different public states will produce the same abstract information set. This may not cause any issues, but we decided to avoid it by training a decoder $\DecoderFunction_\Player: \Abstracted{\Infosets{\Player}} \to \PublicStates$ with loss of L2, which is also propagated through $\AbstractionFunction_\Player$. As a result in each $\Abstracted{\Infoset{\Player}}$ is embedded the information to which public state it belongs. In other words, it also means that information about the public state is also used within $\PropertiesFunction$, but since it is the same for each information set with the same public state, it does not affect the clustering.

\subsection{Off-policy Regularized Nash Dynamics}
The Regularized Nash Dynamics (RNaD) algorithm, which builds upon the Follow the Regularized Leader paradigm, typically updates a player's strategy based on an advantage function \cite{poincare2021}. In the successful Stratego implementation (DeepNash), the authors adapted this by training an information set value function,  which returns a scalar value for an information set. This function approximates the expected game outcome if all players follow the current network strategy in the whole game (both before and after this decision point). This value was then used to derive an advantage function for policy updates, maintaining convergence guarantees within their on-policy training framework.

A strictly on-policy approach may suffer from insufficient exploration, potentially leaving some parts of the game state unvisited. As \AlgorithmName{} aims to find an abstraction in each part of the game, we employed an off-policy sampling strategy. Specifically, we used $\epsilon$-on-policy sampling: at each decision point, an action was chosen uniformly at random with probability $\epsilon$, or sampled from the current network strategy with probability $1 - \epsilon$. This ensures that all parts of the game can be visited with non-zero probability.

However, this off-policy sampling introduces a challenge: a value function trained naively would estimate the value of the $\epsilon$-on-policy sampling strategy, rather than the target network strategy. To obtain an unbiased estimator of the network strategy we trained a history value function. This value function was trained using targets derived from the V-trace off-policy estimator \cite{impala2018}, which corrects for the discrepancy between the sampling policy and the target policy. The advantage  for taking an action at information set was then computed similarly to how RNaD derives advantages from an information set value function \cite{vrmccfr2019, stratego2022}.  Specifically, the advantage relies on counterfactual values - the expected outcome if a specific action is taken and then the network strategy is followed, weighted by the probability opponent have played to this decision point. To ensure unbiased estimates of these counterfactual values under our off-policy sampling scheme, the estimation of this value uses a importance sampling correction for each opponent decision preceding this value \cite{vrmccfr2019,perturbation2025}. 

\subsection{Dynamics network}
n our experiments, we worked with games exhibiting a specific property: the public observation (and thus the public state) at the next step depends only on the previous public state and the joint action taken by all players. In other words, the transition of the public state is independent of the private information distinguishing different information sets within the same current public state.

We leveraged this property to refine our dynamics network ($\DynamicsFunction_\NeuralParameters$). Instead of directly predicting the subsequent abstract information set representation for each player, the modified dynamics function performs a two-stage prediction:
\begin{enumerate}
    \item Predicts the next public state identifier ($\PublicState'$) within the original game.
    \item For each player $\Player$, it predicts a probability distribution over the $\AbstractionLimit$ abstract information set within that predicted next public state $\PublicState'$.
\end{enumerate}

To determine the actual subsequent abstract information set representation for player $\Player$, we first use public state representation function $\AbstractionFunction_{\Player, \NeuralParameters}(\PublicState')$ to get the $\AbstractionLimit$ abstract information sets. Then we select the $\Abstracted{\Infoset{\Player}}$, which corresponds to the highest probability from the dynamics network (argmax selection). This modification effectively separates the prediction of the public state transition from the prediction of the players' abstracted private states within that future public context. It can simplify the learning task for the dynamics network when the underlying game structure supports this decomposition.

\begin{table}[htbp]
    \centering
    \begin{tabular}{lll}
        \toprule
        \textbf{Group} & \textbf{Parameter} & \textbf{Value} \\
        \midrule
        \multirow{9}{*}{\textbf{RNaD}} 
            & Regularization $\eta$ & 0.2 \\
            & NeuRD threshold $\beta$ & 2 \\
            & NeuRD clipping $c$ & 10000 \\
            & V-trace clipping $c$ & 1.0 \\
            & V-trace clipping $\rho$  & $\infty$ \\ 
            & Target network update $\gamma_a$  & $10^{-3}$\\
            & Off-policy sampling $\epsilon$ & 0.5 \\
            & Amount of transformations $\Transformations$ & 10 \\
        \midrule
        \multirow{8}{*}{\textbf{Training}} 
            & Learning rate & $3\cdot 10^{-4}$\\
            & Architecture & MLP \\
            & Activation functions & ReLU \\
            & Optimizer & ADAM (ADAMW for $\PropertiesFunction_\NeuralParameters$) \\
            & Adam decay rate $\beta_1$ & 0.9 (0 for $\Strategy{\NeuralParameters}$) \\
            & Adam decay rate $\beta_2$ & 0.999\\
            & Weight decay of ADAMW & $10^{-5}$ \\
            & Gradient clipping & 100 (1 for $\PropertiesFunction_\NeuralParameters, \TransformationsFunction_\NeuralParameters$) \\
        \midrule
        \multirow{1}{*}{\textbf{Abstractions}}
            & Sample noise scale $\sigma$ & 0.02 \\
        \midrule
        \multirow{3}{*}{\textbf{Soft clustering}}
            & Softmax temperature $\gamma$ & 1 \\
            & Hard clustering threshold $\HardClusteringDist$ & 0.3 \\
            & Closest cluster distance $\RepulsionConstant$ & 0.5 \\
        \midrule
        \multirow{2}{*}{\textbf{Look-ahead reasoning}}
            & CFR+ iterations & 1000 \\
            & Depth limit & 1 \\
        \bottomrule
    \end{tabular}
    \caption{Common hyperparameters}
    \label{tab:hyperparameters}
\end{table}

\begin{table}[htbp]
    \centering
    \begin{tabular}{l|ccc}
        \toprule
        \textbf{Parameter} & \textbf{Exploitability} & \textbf{Leduc} & \textbf{Head-to-head} \\
        \midrule
        \multicolumn{4}{l}{\textit{Architecture Parameters}} \\
        \midrule
        $\Strategy{\NeuralParameters}$ MLP layers & 256, 256 & 256, 256 & 512, 512 \\ 
        $\ValueFunction_\NeuralParameters$ MLP layers & 512, 512 & 512, 512 & 4096, 4096 \\ 
        $\RepresentationFunction_\NeuralParameters$ MLP layers & 256, 256 & 256, 256 & 512, 512 \\ 
        $\AbstractionFunction_\NeuralParameters$ MLP layers & 1024, 1024 & 256, 256 & 4096, 4096 \\ 
        $\LegalActionsFunction_\NeuralParameters$ MLP layers & 128, 128 & 128, 128 & 512, 512 \\ 
        $\DynamicsFunction_\NeuralParameters$ MLP layers  & 256, 256 & 256, 256 & 2048, 2048 \\ 
        $\PropertiesFunction_\NeuralParameters$ MLP layers  & 256, 256 & 256, 256 & 1024, 1024 \\ 
        $\TransformationsFunction_\NeuralParameters$ MLP layers  & 512, 512 & 512, 512 & 4096, 4096 \\ 
        $\DecoderFunction_\NeuralParameters$ MLP layers  & 128, 128 & 128, 128 & 512, 512 \\ 
        \midrule
        \multicolumn{4}{l}{\textit{Training Parameters}} \\
        \midrule
        Batch size & 64 & 128 & 128 \\ 
        Episodes & $10^5$ & $10^5$ & $3\cdot10^6$ \\
        \midrule
        \multicolumn{4}{l}{\textit{Other Parameters}} \\
        \midrule
        Abstract information set dimension  & 64 & 64 & 256\\
        Regularization policy change each & 1000 & 1000 & 20000 \\
        \bottomrule
    \end{tabular}
    \caption{Specific hyperparameters for each experiment.}
    \label{tab:hyperparameters_experiment}
\end{table}

\section{Additional experimental details}

\subsection{Tabular K-means}
\label{sec:tabular_experiment}  
\begin{figure}[h]
    \centering
    \begin{subfigure}[b]{0.3\textwidth}
        \centering
        \includegraphics[width=\textwidth]{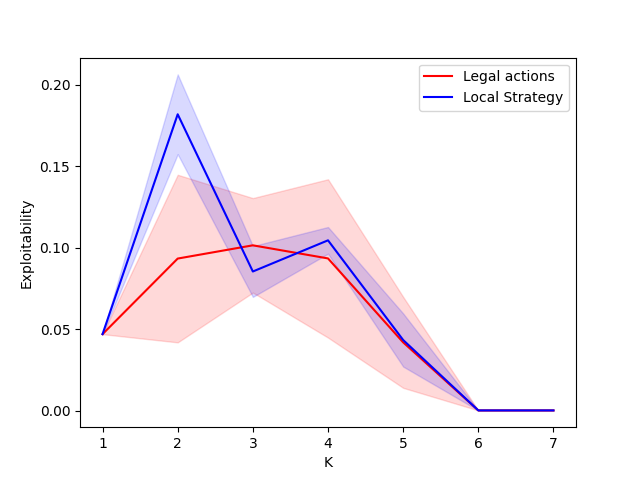}
        \caption{Goofspiel 4}
        \label{fig:kmeans_goof4}
    \end{subfigure}
    \hfill  
    \begin{subfigure}[b]{0.3\textwidth}
        \centering
        \includegraphics[width=\textwidth]{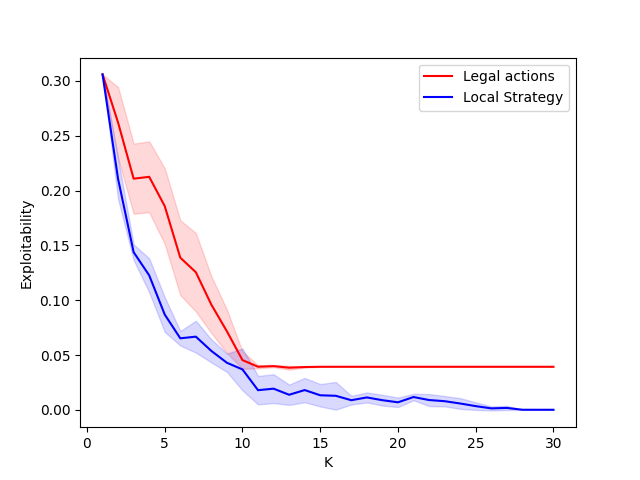}
        \caption{Goofspiel 5}
        \label{fig:kmeans_goof5}
    \end{subfigure}
    \hfill 
    \begin{subfigure}[b]{0.3\textwidth}
        \centering
        \includegraphics[width=\textwidth]{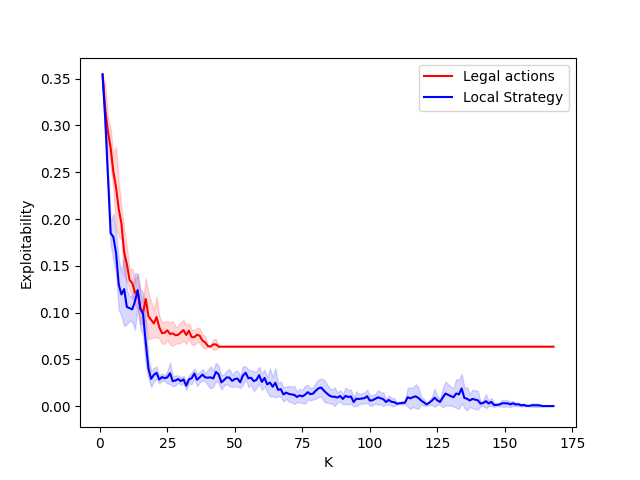}
        \caption{Goofspiel 6}
        \label{fig:kmeans_goof6}
    \end{subfigure}
    \caption{Exploitability in different games when performing tabular K-means to get abstraction using different property functions $\PropertiesFunction$.}
    \label{fig:tabular_experiment}
\end{figure}

In this experiment, we evaluate abstraction performance separately to show that tabular clustering of the real information sets in each public state to a limited number of abstract ones leads to graceful degradation of the quality of the strategies. As a clustering property $\PropertiesFunction$, we have used either legal action or strategy computed by 4,000 iterations of Counterfactual Regret Minimization. We performed the tabular K-means as a clustering algorithm and then we constructed the original game tree, while changing the information structure to use the abstraction. This new abstract game was solved using CFR and then the exploitability of this final strategy was computed in the original game. The results are displayed in \cref{fig:tabular_experiment}.

The maximal amount of information sets per player in public state is 7, 30 and 168 for Goofspiel 4, 5 and 6 respectively. Increasing the abstraction limit beyond 10 in Goofspiel 5 and 20 in Goofspiel 6 increases the performance only slightly, so it suggests that this size is sufficient in those games. We have used this knowledge in our large experiments and we set the abstraction limit $\AbstractionLimit = 20$. Also, it is important to note that legal actions do not provide sufficient information to create an optimal abstraction in larger Goofspiels. However, in large games, as tested in \cref{sec:heads_experiment}, it still outperformed the RNaDs strategy as $\PropertiesFunction$. 
\paragraph{Computational resources:} For each value of $K$ in the experiment, we have used a single core of AMD EPYC 7543 for all 10 seeds. For each $K$ in Goofspiel 4, the computation took less than a minute, in Goofspiel 5 it took less than hour and in Goofspiel 6 it was less than 8 hours. However, all of those were ran in parallel. Approximately the resulting computational time was at most 1375 CPU hours.

\subsection{Exploitability in Smaller Games}
We provide additional experimental results for \cref{sec:exploitability_experiment}, which were omitted from the main body of the paper due to limited space. We show a performance of different $\PropertiesFunction$ from those in the main body and also different ablation studies when computing exploitability from the same training runs.

\begin{figure}[htbp]
    \centering
    
    \begin{minipage}[c]{0.08\textwidth}
        \centering
        \rotatebox{90}{\textbf{Abstraction w/o dynamics}}
    \end{minipage}%
    \begin{minipage}[c]{0.92\textwidth} 
        \begin{subfigure}[b]{0.49\textwidth}
            \centering
            \includegraphics[width=\textwidth]{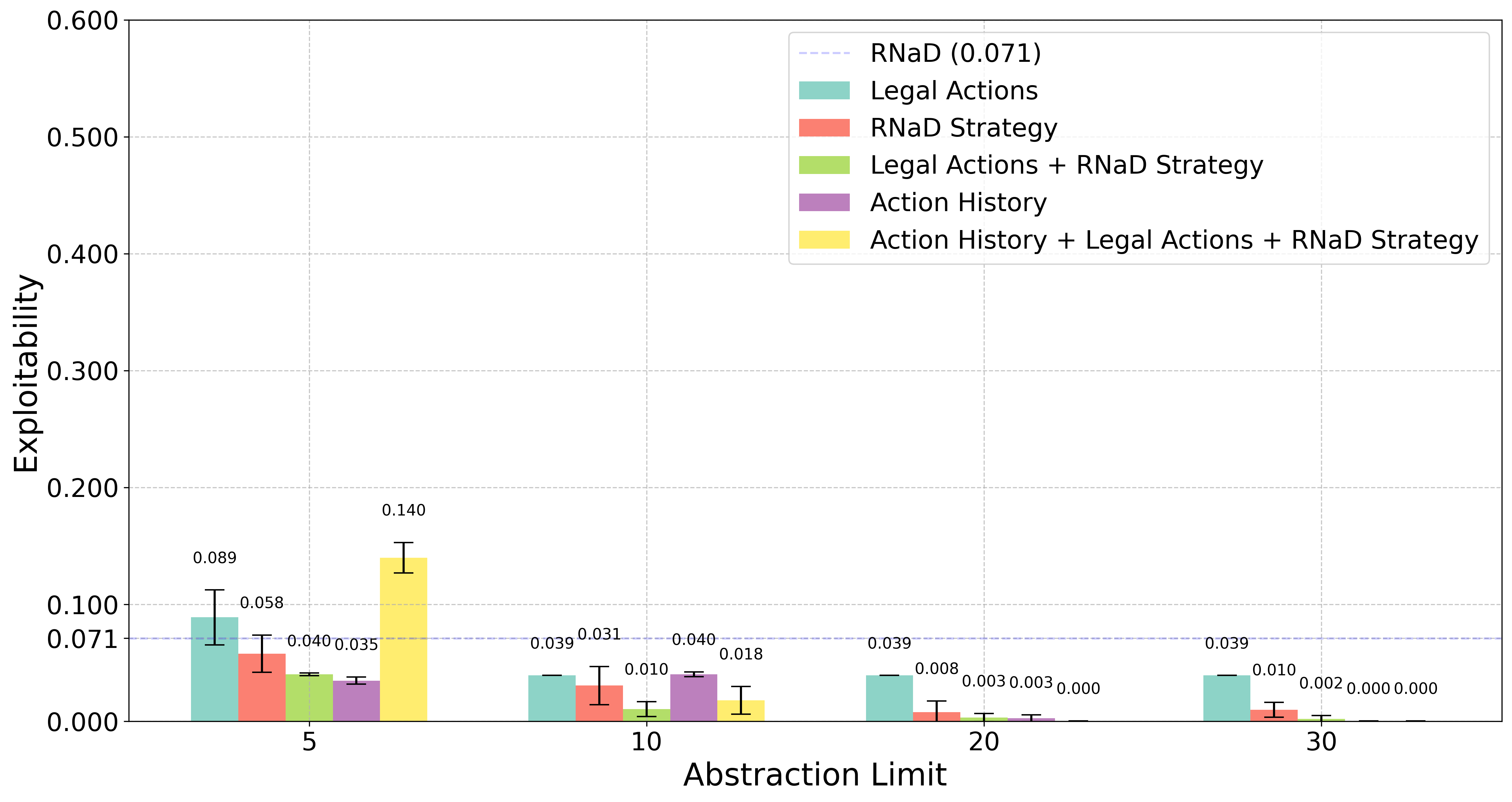}
            \label{fig:app_goofspiel_5_abs}
        \end{subfigure}%
        \hfill
        \begin{subfigure}[b]{0.49\textwidth}
            \centering
            \includegraphics[width=\textwidth]{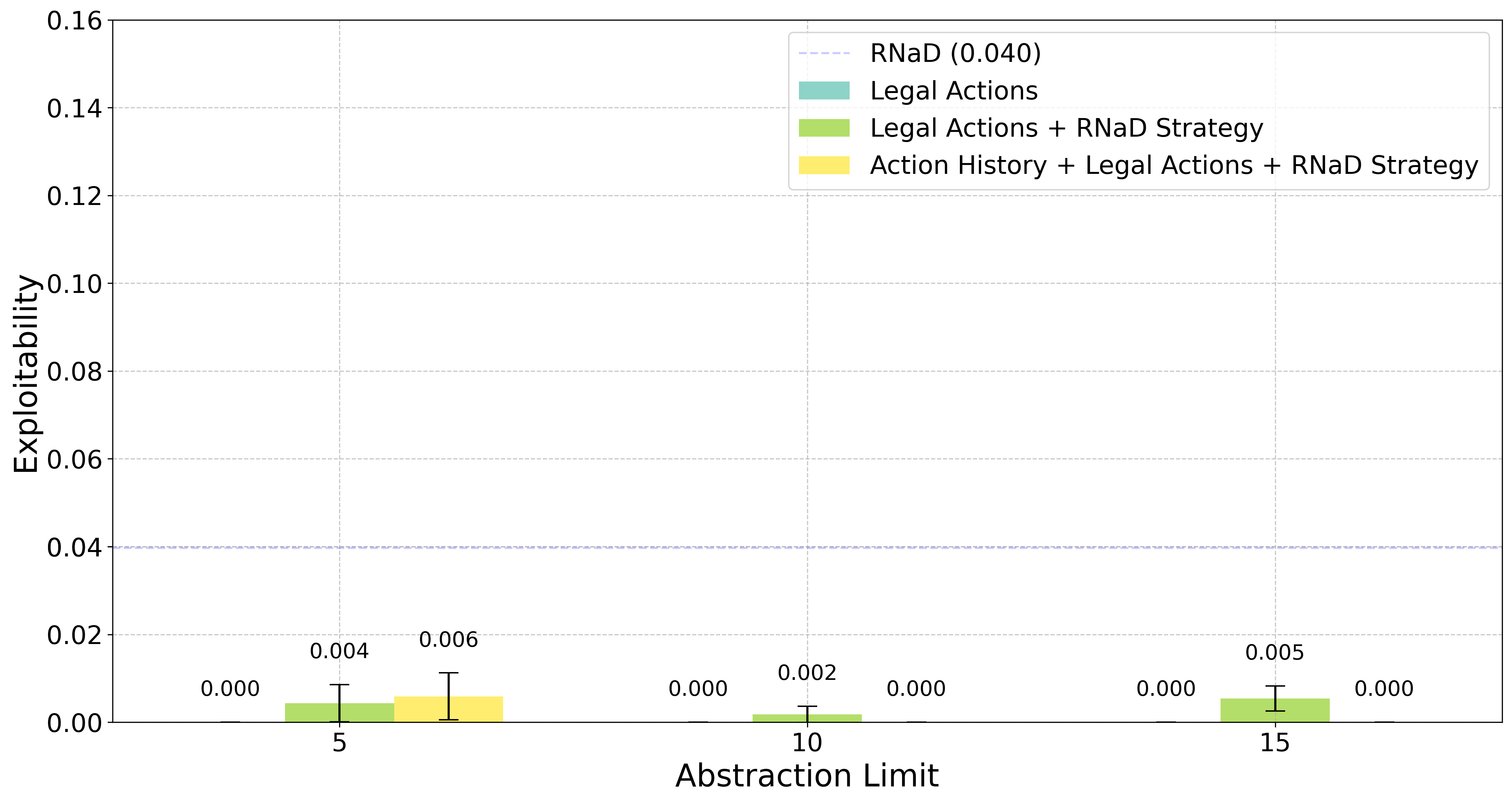}
            \label{fig:app_oshi_7_5_abs}
        \end{subfigure}
    \end{minipage}
    \begin{minipage}[c]{0.08\textwidth}
        \centering
        \rotatebox{90}{\textbf{Full game}}
    \end{minipage}%
    \begin{minipage}[c]{0.92\textwidth} 
        \begin{subfigure}[b]{0.49\textwidth}
            \centering
            \includegraphics[width=\textwidth]{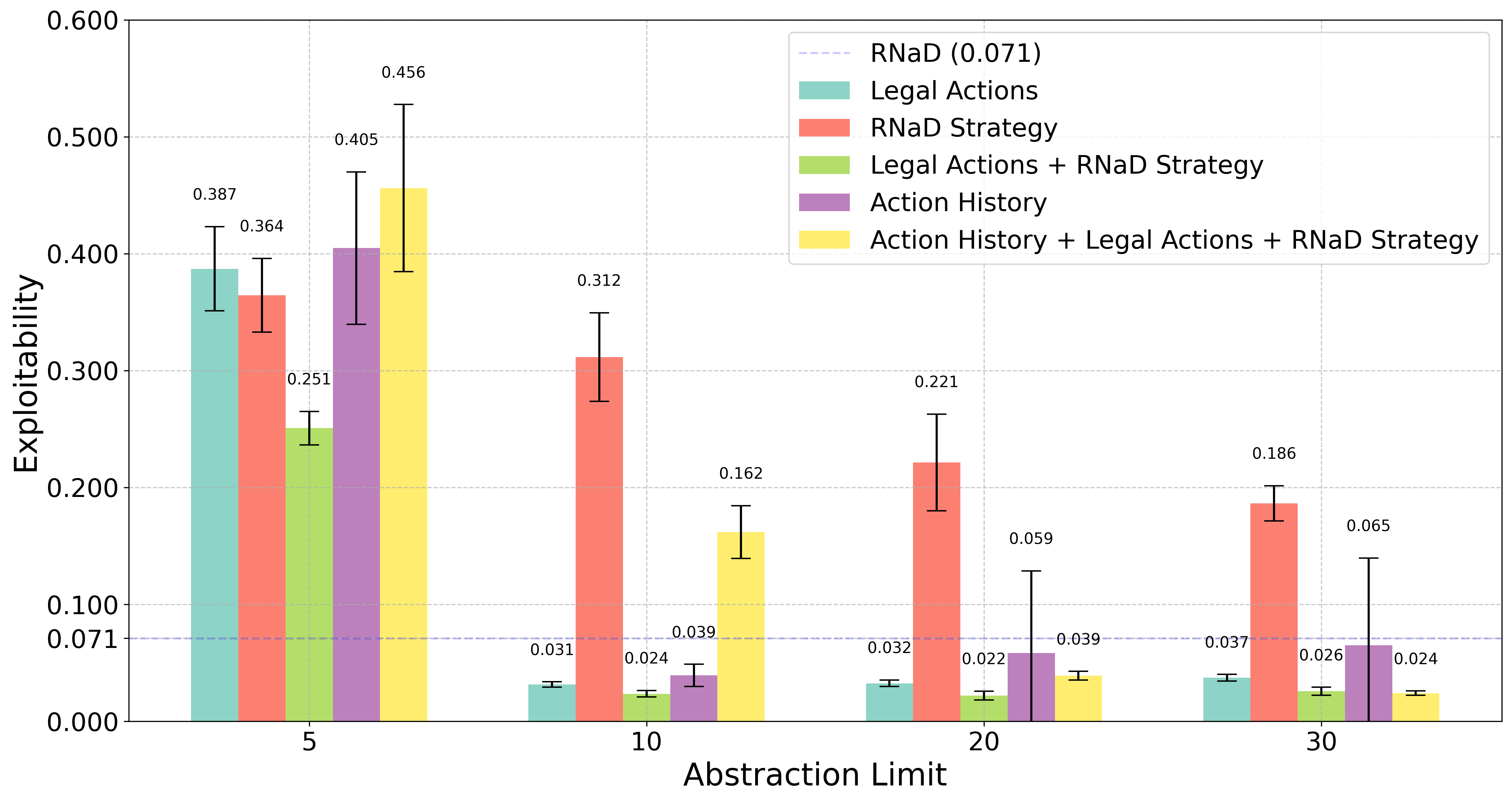}
            \label{fig:app_goofspiel_5_full}
        \end{subfigure}%
        \hfill
        \begin{subfigure}[b]{0.49\textwidth}
            \centering
            \includegraphics[width=\textwidth]{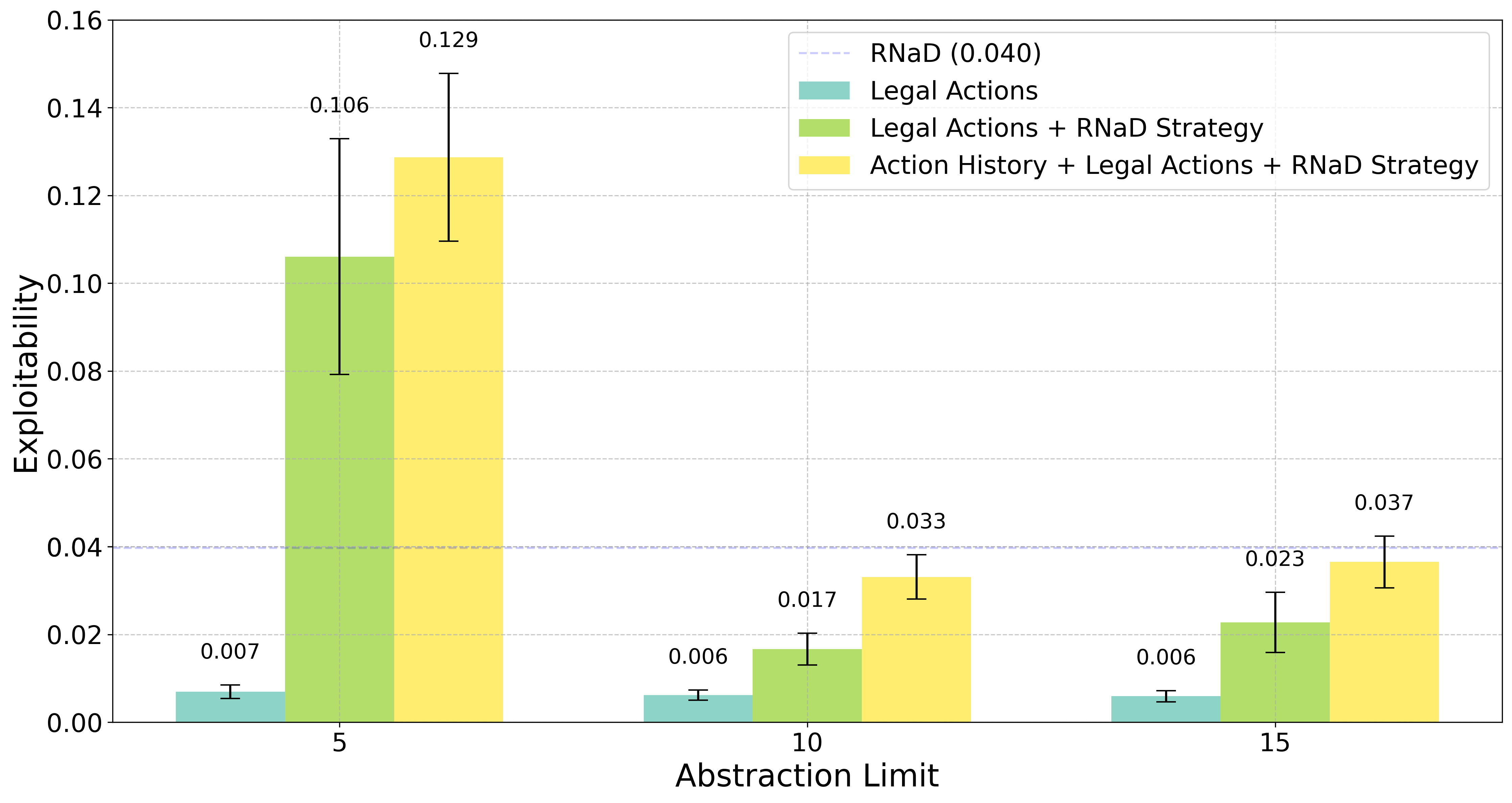}
            \label{fig:app_oshi_zumo_7_5_full}
        \end{subfigure}
    \end{minipage}
    \begin{minipage}[c]{0.08\textwidth}
        \centering
        \rotatebox{90}{\textbf{\AlgorithmName{}}}
    \end{minipage}%
    \begin{minipage}[c]{0.92\textwidth} 
        \begin{subfigure}[b]{0.49\textwidth}
            \centering
            \includegraphics[width=\textwidth]{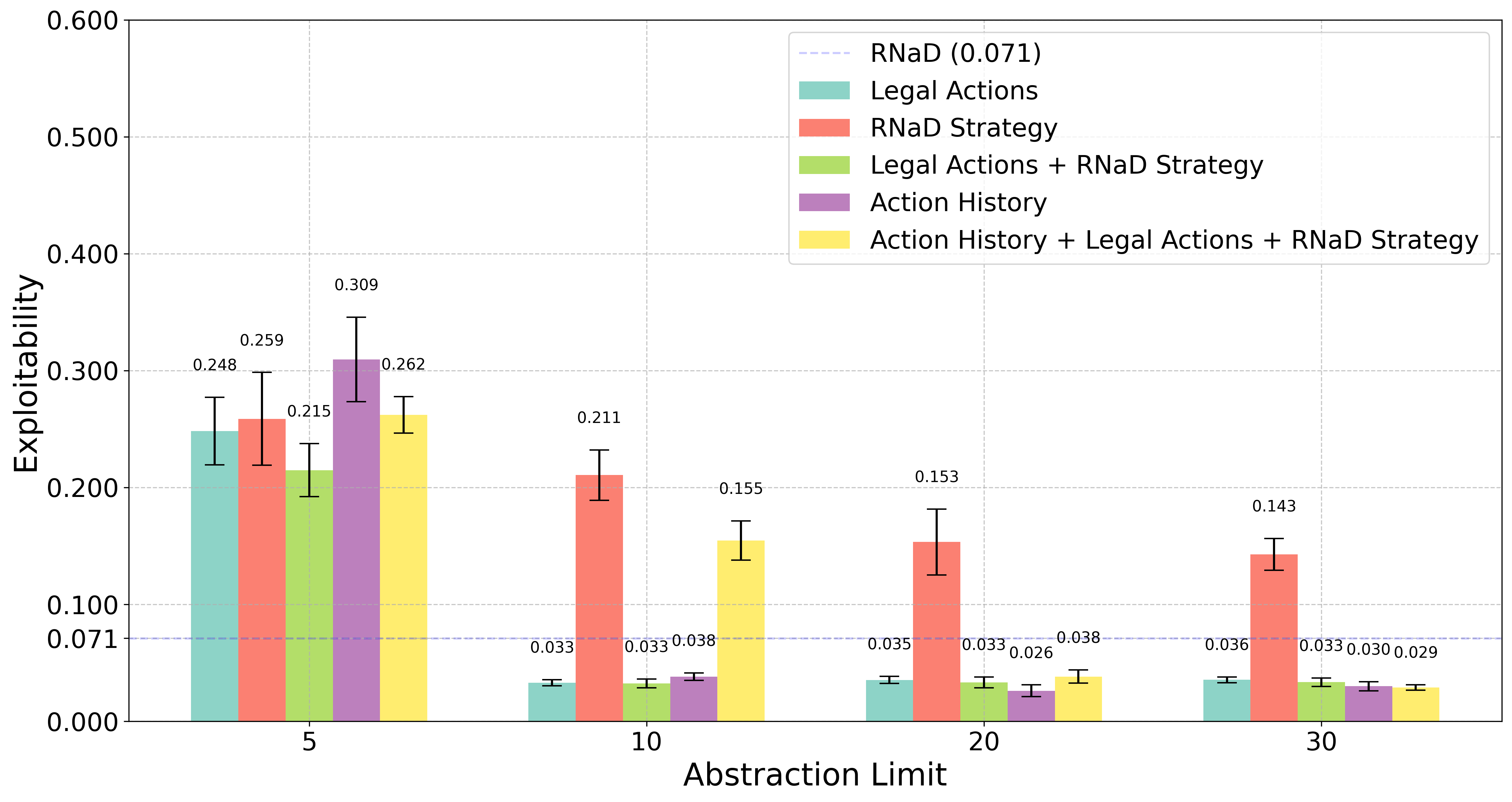}
            \captionsetup{labelformat=empty}
            \caption{Goofspiel 5}
            \label{fig:app_goofspiel_5_subgame}
        \end{subfigure}%
        \hfill
        \begin{subfigure}[b]{0.49\textwidth}
            \centering
            \includegraphics[width=\textwidth]{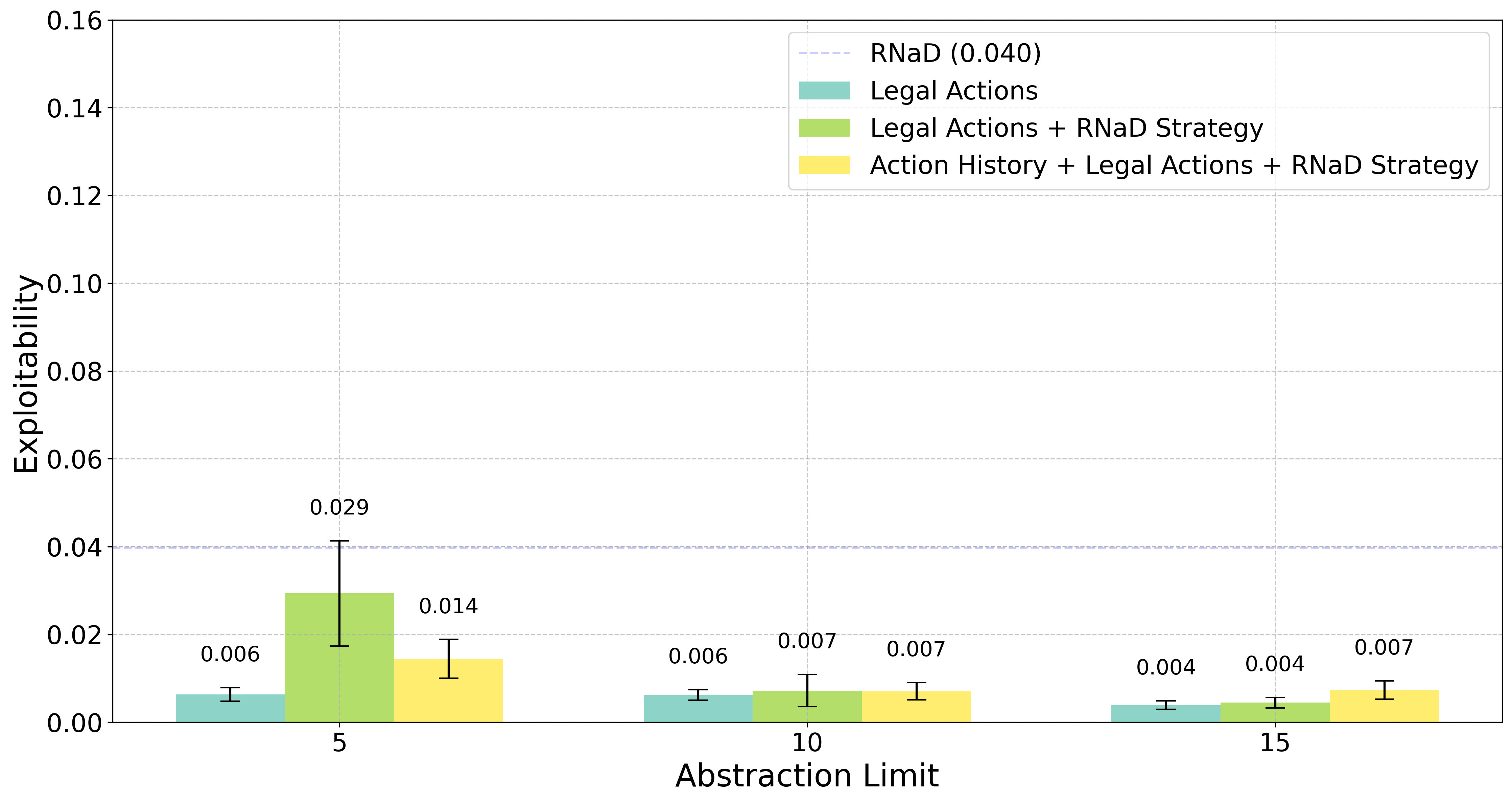}
            \captionsetup{labelformat=empty}
            \caption{Oshi-Zumo 3,5}
            \label{fig:app_oshi_zumo_7_5_subgame}
        \end{subfigure}
    \end{minipage}
    \caption{Exploitability of different \AlgorithmName{} runs  with different $\PropertiesFunction$ either by mapping the information abstraction onto the original game tree, or by constructing the whole game tree from the dynamics network or by using the \AlgorithmName{} with depth-limit 1}
    \label{fig:app_exploitability}
\end{figure}

\cref{fig:app_exploitability} contains the exploitability of the abstraction in three different settings. First, we construct the original game tree, and then we map each infoset with public state representation function $\AbstractionFunction_\NeuralParameters$ and information set representation $\RepresentationFunction_\NeuralParameters$ to the abstract one. Then, we solved this changed game with CFR+ and computed the exploitability of this new strategy in the original one. The Second was to use a dynamics network to construct the whole game, which is equivalent to using $\infty$ depth-limit and then using CFR+ in this game. This setting does not use a value function, so it evaluates the quality of the abstraction $\AbstractionFunction_\NeuralParameters$ and the dynamics $\DynamicsFunction_\NeuralParameters$. Third, we used the \AlgorithmName{}, which is the same setting as in \cref{sec:abstraction}.

In Goofspiel 5, the experiments have shown that using just the RNaD strategy $\StrategyFunction_\NeuralParameters$, which was trained for the $\ValueFunction_\NeuralParameters$ for clustering $\PropertiesFunction$ performs worse than the RNaD itself. This is partially due to changes to the clustering to avoid collapsing clusters together, which we discussed in \cref{app:soft_cluster} and also because the strategy itself was not stationary during the training, as it changed from the RNaD dynamics. 

Using just the abstraction and then mapping it to the original game shows that with sufficient capacity, the learned abstraction mirrors the game's underlying structure, as evidenced by $\AbstractionLimit=30$ in Goofspiel. In Oshi-Zumo, even with $\AbstractionLimit \geq 5$ it is enough to model the game perfectly when using the legal actions for clustering $\PropertiesFunction$. This suggests that even if Oshi-Zumo is quite a large game, as the largest public state contains 625 information sets, it is not important to distinguish between them.

Constructing the whole game tree sometimes produces worse results than using the continual resolving. This mainly happens if the abstraction is worse than the value function, which may fix some mistakes that the poor abstraction caused only further in the game. This mainly happened in Oshi-Zumo. In goofspiel it occurred only when using only the action history as $\PropertiesFunction$.

\paragraph{Computational resources:} Training of each seed, abstraction limit $\AbstractionLimit$, clustering property $\PropertiesFunction$ and game was ran on a single GPU Nvidia Tesla A100 for less than an hour. Then each checkpoint made during this training was evaluated sequentially on a single CPU AMD EPYC 7543 for each ablation study. So in total this experiment cost 390 GPU hours and 6240 CPU hours.

\subsection{Leduc Hold'em}
\label{sec:leduc_experiment}
\begin{figure}
    \centering
    \includegraphics[width=0.5\linewidth]{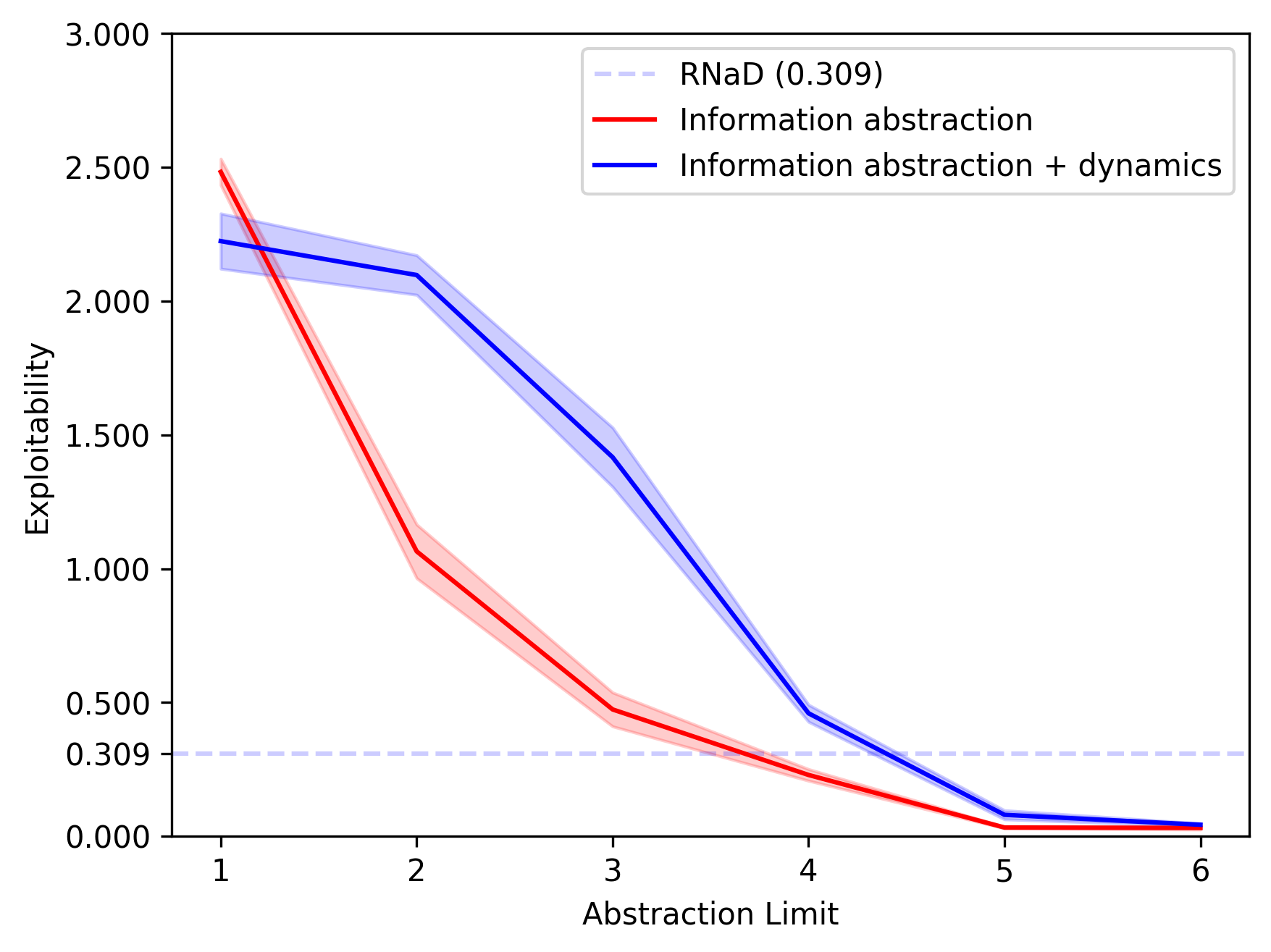}
    \caption{Exploitability in Leduc Hold'em either by constructing the subgame from the rules of the game, or from the dynamics network}
    \label{fig:leduc_exploitability}
\end{figure}

We have also evaluated \AlgorithmName{} in a small version of Poker, Leduc Hold'em, which is a popular benchmark. \AlgorithmName{} cannot model the chance nodes and is not intended to be used for such games. However, with some domain-specific workarounds, we were able to use \AlgorithmName{} even for Leduc. These workarounds were only used in the test part to unroll the chance nodes out of game rules of Leduc. We have used the same training setting as in \cref{sec:exploitability_experiment} with 100,000 episodes and evaluated the last 21 checkpoints each 1000 episodes. We evaluated the same trained models in two settings. In both we solved separately part of the game before dealing a public card, by using a value function at the depth-limit and after dealing a public card. In one setting, we have created the subgame from the rules of the game and only mapped the information structure from the abstraction; in the second, we only used the rules of the game to construct states after the chance nodes and then used the dynamics network to construct the rest of the subgames. The results are displayed in \cref{fig:leduc_exploitability}.

In Leduc Hold'em, each public state has 6 information sets for each player. So for $\AbstractionLimit \geq 6$, \AlgorithmName{} learns the underlying game. However, even with $\AbstractionLimit = 5$, it is still capable of outperforming RNaD. Decreasing the size further degrades the performance, even if the abstraction $\AbstractionLimit = 3$ should be sufficient due to the invariance in the card suits. This suggests that improving $\PropertiesFunction$ and the clustering may produce better results.

\paragraph{Computational resources:} We have again used for a training of a single seed with abstraciton limit just a single GPU Nvidia Tesla A100 for at most hour. Each checkpoint was then evaluated on a single CPU AMD EPYC 7543 for each of those 2 ablation studies in less than 2 hours. In total, the computational cost was 60 GPU hours and 120 CPU hours.

\subsection{Head-to-head in Larger Games}
\paragraph{Computational resources:} The training of each seed and clustering function $\PropertiesFunction$ for both RNaD and \AlgorithmName{} was done in parallel each on a single GPU Nvidia Tesla A100. The training took 24 hours. Heads-to-heads was done performed in parallel for each final saved model from training. The evaluation of each pair took 192 hours on a CPU AMD EPYC 7543. The total computational cost was 864 GPU hours and 20736 CPU hours.

\section{Game rules}
\label{sec:app_game_rules}
\subsection{Leduc Hold'em}
Leduc Hold'em is a simplified version of Texas Hold'em poker. It is played by two players with a deck of 6 cards in two suits: Spades and Hearts. Each suit has three ranks: Jack, Queen, and King. At the beginning of the game, each player performs a mandatory bet of 1 coin to the pot, and the dealer deals privately one card to each player. The game then proceeds to the first betting round, where players take turns by either folding, calling, or raising. If some player folds, the game immediately ends, and the other player receives all the coins in the pot. If any player calls, it puts as many coins into the pot so that the total amount of coins put in by both players is the same, and the game proceeds to the next round. If a player raises, it puts the same amount of coins and two more coins into the pot as during the call. The players can raise only twice during a single round. After the first round ends, one of the remaining four cards is revealed as public. Then, the players proceed with the betting round. The only difference is that the raise now adds four more coins to the pot. At the end of this betting round, both players reveal the following rule that decides their card and the winner: If a player's private card matches the rank of the public card (e.g., the player has a King and the public card is a King), they have a pair and win. Otherwise, the player holding a higher value card wins, and the order from the highest rank is King, Queen, Jack. If both players hold the same rank, it is a draw, and they will split the money in the pot.

\subsection{Imperfect Information Goofspiel}
Imperfect information Goofspiel $N$ is a game played by two players, where each player receives cards valued from 1 to $N$. The dealer has the same cards. Then each turn, the dealer reveals a single card from its deck to both players. Each player then secretly places one of its cards as a bet. The dealer looks at both cards and gives the points corresponding to the public card to the player that had the highest bid. In case of a draw, the dealer discards its card. The players can only play the same card once. We have used a version where the dealer has predefined order of cards, so it always shows from the highest card to the lowest one.

\subsection{Imperfect Information Oshi-Zumo}
Imperfect information Oshi-Zumo $K,N$ is played by two players on a board of size $2K + 1$ with a fighter in the middle of the board. Each player starts with $N$ coins. Then each turn, players secretly place bids from 0 to the amount of coins they are still holding. The player that had the higher bet pushes the fighter closer to the opponents edge. The game ends either when neither of the players has any coins, when the fighter is on the edge of the board (positions 0 and $2K$) or after the maximum number of rounds. We have used $N$ as the number of rounds. The reward of player 1 is then $\Rewards{1} = \frac{P - K}{K}$, where $P$ is the position of the fighter at the end of the game.

\section{Large Language Models usage}
During the writing of this paper, the Large Language Models (LLMs) were used to refine the writing, both by polishing the text and to better communicate the main contributions of the paper. Specifically, we have used Gemini 2.0 Flash, ChatGPT, and Claude Sonnet 4.

We have also used the LLM coding assistant Cursor with Claude Sonnet 3.5 and later Claude Sonnet 4 as the underlying model for the experimental evaluation.

The authors always double-checked all of the LLM outputs to ensure their correctness.

\end{document}